\documentclass[lettersize,journal]{IEEEtran}
\usepackage{amsmath,amsfonts}
\usepackage{algorithmic}
\usepackage{algorithm}
\usepackage{array}
\usepackage[caption=false,font=normalsize,labelfont=sf,textfont=sf]{subfig}

\usepackage{graphicx}
\usepackage{amsmath}
\usepackage{amssymb}
\usepackage{booktabs}

\usepackage{algorithm}
\usepackage{algorithmic}

\usepackage{epsfig}
\usepackage{enumitem}
\usepackage{bbding} 

\usepackage{times}
\usepackage{caption}
\usepackage{subcaption}
\usepackage{array}
\usepackage{colortbl}
\usepackage{bbm}
\usepackage{multirow}
\usepackage{multicol}
\usepackage{makecell}
\usepackage{xcolor}
\usepackage{xspace}
\usepackage{color}

\usepackage{pifont}

\usepackage{tabulary}
\usepackage{dblfloatfix}
\usepackage{transparent}

\usepackage{algorithm}
\usepackage{listings}
\usepackage{algorithmic}
\usepackage{balance}

\usepackage{color, colortbl}
\definecolor{iGray}{gray}{0.9}
\definecolor{beaublue}{rgb}{0.74, 0.83, 0.9}
\definecolor{Royal_Blue}{rgb}{0.0, 0.1, 0.66}
\definecolor{cGreen}{RGB}{100,180,100}
\usepackage[pagebackref=true,breaklinks=true,letterpaper=true,colorlinks,bookmarks=false, citecolor = Royal_Blue]{hyperref}

\usepackage{tabulary}
\usepackage{dblfloatfix}
\usepackage{transparent}
\usepackage{threeparttable}
\usepackage{longtable}
\usepackage{rotating}
\usepackage{tabularx}
\usepackage{mwe}
\usepackage{bm}
\usepackage{float}

\newlength\savewidth
\newcommand{\tablestyle}[2]{\setlength{\tabcolsep}{#1}\renewcommand{\arraystretch}{#2}\centering\footnotesize}
\renewcommand{\paragraph}[1]{\vspace{.5mm}\noindent\textbf{#1}}

\newcolumntype{x}[1]{>{\centering\arraybackslash}p{#1pt}}
\newcolumntype{y}[1]{>{\raggedright\arraybackslash}p{#1pt}}
\newcolumntype{z}[1]{>{\raggedleft\arraybackslash}p{#1pt}}

\newcommand{\app}{\raise.17ex\hbox{$\scriptstyle\sim$}}

\definecolor{deemph}{gray}{0.6}

\definecolor{baselinecolor}{gray}{.9}

\usepackage{pifont}
\usepackage{bbding}
\usepackage{threeparttable}
\usepackage{multirow}
\usepackage{longtable}
\usepackage{rotating}
\usepackage{tabularx}
\usepackage{graphicx}
\usepackage{mwe}
\usepackage{amsmath, bm}
\usepackage{float}
\usepackage{caption}
\usepackage{pifont}
\usepackage{bbding}
\usepackage{graphicx}
\usepackage{mwe}
\usepackage{amsmath, bm}
\usepackage{float}
\usepackage{diagbox}
\usepackage{caption}
\usepackage{graphicx}
\usepackage{amsfonts}
\usepackage{color}
\usepackage{bm}
\usepackage{amsmath}
\usepackage{amssymb}
\usepackage{multirow}
\usepackage{makecell}
\usepackage{enumitem}
\usepackage{tabularx}
\usepackage{fixltx2e} 
\usepackage{booktabs}
\usepackage{xcolor}
\usepackage{url}
\usepackage{graphicx}
\usepackage{amsfonts}
\usepackage{color}
\usepackage{bm}
\usepackage[capitalize]{cleveref}
\crefname{section}{Sec.}{Secs.}
\Crefname{section}{Section}{Sections}
\Crefname{table}{Table}{Tables}
\crefname{table}{Tab.}{Tabs.}

\ifCLASSINFOpdf
\else

\fi
\begin{document}

\title{Correlation-Embedded Transformer Tracking: \\ A Single-Branch Framework}

\author{Fei Xie,
        Wankou Yang,
        Chunyu Wang,
        Lei Chu, 
        Yue Cao,
       Chao Ma,
       Wenjun Zeng,~\IEEEmembership{Fellow,~IEEE}
\thanks{

Received 17 January 2024; revised 8 June 2024, accepted 10 August 2024. This work was supported in part by NSFC under Grant 62322113 and Grant62376156, in part by Shanghai Municipal Science and Technology Major Projectunder Grant 2021SHZDZX0102, and in part by the Fundamental Research Fundsfor the Central Universities. Corresponding author: Chao Ma.

 F. Xie and C. Ma are with the MoE Key Lab of Artificial Intelligence, AI Institute, Shanghai Jiao Tong University, Shanghai, China. E-mail: \{jaffe031, chaoma\}@sjtu.edu.cn.

 W. Yang is with the School of Automation, Southeast University, Nanjing, China. E-mail: wkyang@seu.edu.cn.

 C. Wang, L. Chu, and Y. Cao are with Microsoft Research Asia, Beijing, China. E-mail: \{chnuwa, leichu\}@microsoft.com, caoyue10@gmail.com.

 W. Zeng is with the Eastern Institute of Technology, Ningbo, China. E-mail: wenjunzengvp@eias.ac.cn.

Code is available at \url{https://github.com/VISION-SJTU/SuperSBT}.

This article has supplementary downloadable material available at
https://doi.org/10.1109/TPAMI.2024.3448254, provided by the authors

}
}



\maketitle

\begin{abstract}
Developing robust and discriminative appearance models has been a long-standing research challenge in visual object tracking. In the prevalent Siamese-based paradigm, the features extracted by the Siamese-like networks are often insufficient to model the tracked targets and distractor objects, thereby hindering them from being robust and discriminative simultaneously. While most Siamese trackers focus on designing robust correlation operations, we propose a novel single-branch tracking framework inspired by the transformer. Unlike the Siamese-like feature extraction, our tracker deeply embeds cross-image feature correlation in multiple layers of the feature network. By extensively matching the features of the two images through multiple layers, it can suppress non-target features, resulting in target-aware feature extraction. The output features can be directly used to predict target locations without additional correlation steps. Thus, we reformulate the two-branch Siamese tracking as a conceptually simple, fully transformer-based Single-Branch Tracking pipeline, dubbed SBT. 
After conducting an in-depth analysis of the SBT baseline, we
summarize many effective design principles and propose an improved tracker dubbed SuperSBT.
SuperSBT adopts a hierarchical architecture with a local modeling layer to enhance shallow-level features.  
A unified relation modeling is proposed to remove complex handcrafted layer pattern designs. 
SuperSBT is further improved by masked image modeling pre-training, integrating temporal modeling, and equipping with dedicated prediction heads.
Thus, SuperSBT outperforms the SBT baseline by $4.7$\%,$3.0$\%, and $4.5$\% AUC scores in LaSOT, TrackingNet, and GOT-10K. 
Notably, SuperSBT greatly raises the speed of SBT from $37$ FPS to $81$ FPS. 
Extensive experiments show that our method achieves superior results on eight VOT benchmarks. 

\end{abstract}


\IEEEdisplaynontitleabstractindextext

\IEEEpeerreviewmaketitle

\section{Introduction}\label{sec:introduction}

\IEEEPARstart{V}{isual} object tracking (VOT) is a well-established subject in the field of computer vision.
In recent years, there have been notable advancements in the visual tracking area. Nonetheless, it continues to pose a significant challenge, especially in real-world scenarios, due to various factors, such as changes in lighting and object size, complex background clutter, and obstructions.
Object tracking poses a significant challenge due to the fundamental yet competing goals it needs to achieve. On the one hand, it needs to recognize the target despite its changing appearance. On the other hand, it needs to filter out distractor objects in the background that may be very similar to the target.%

\begin{figure}[t]
	\centering{\includegraphics[scale = 0.55]{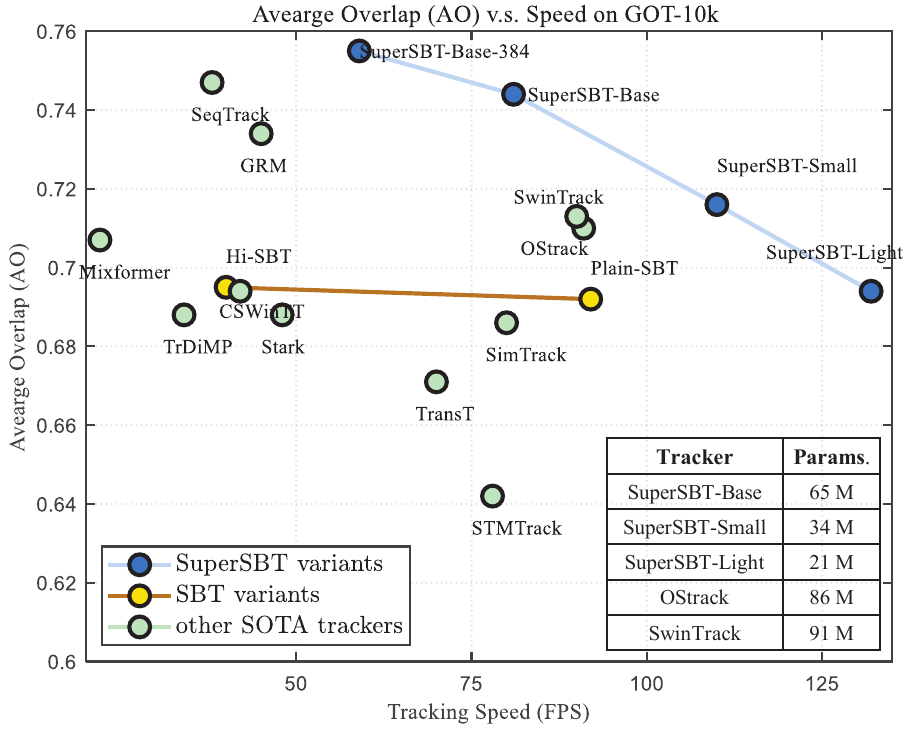}} 
	\caption{Comparison of state-of-the-art trackers on GOT-10k~\cite{got}. We visualize the AO performance with respect to the model size and running speed. All reported results follow the official GOT-10k test protocol. Our SBT and SuperSBT variants achieve superior results with high speed.} 
	\label{fig:bab}
\end{figure}

\begin{figure*}
\begin{minipage}[t]{0.50\linewidth}
\centering
\includegraphics[scale = 0.26]{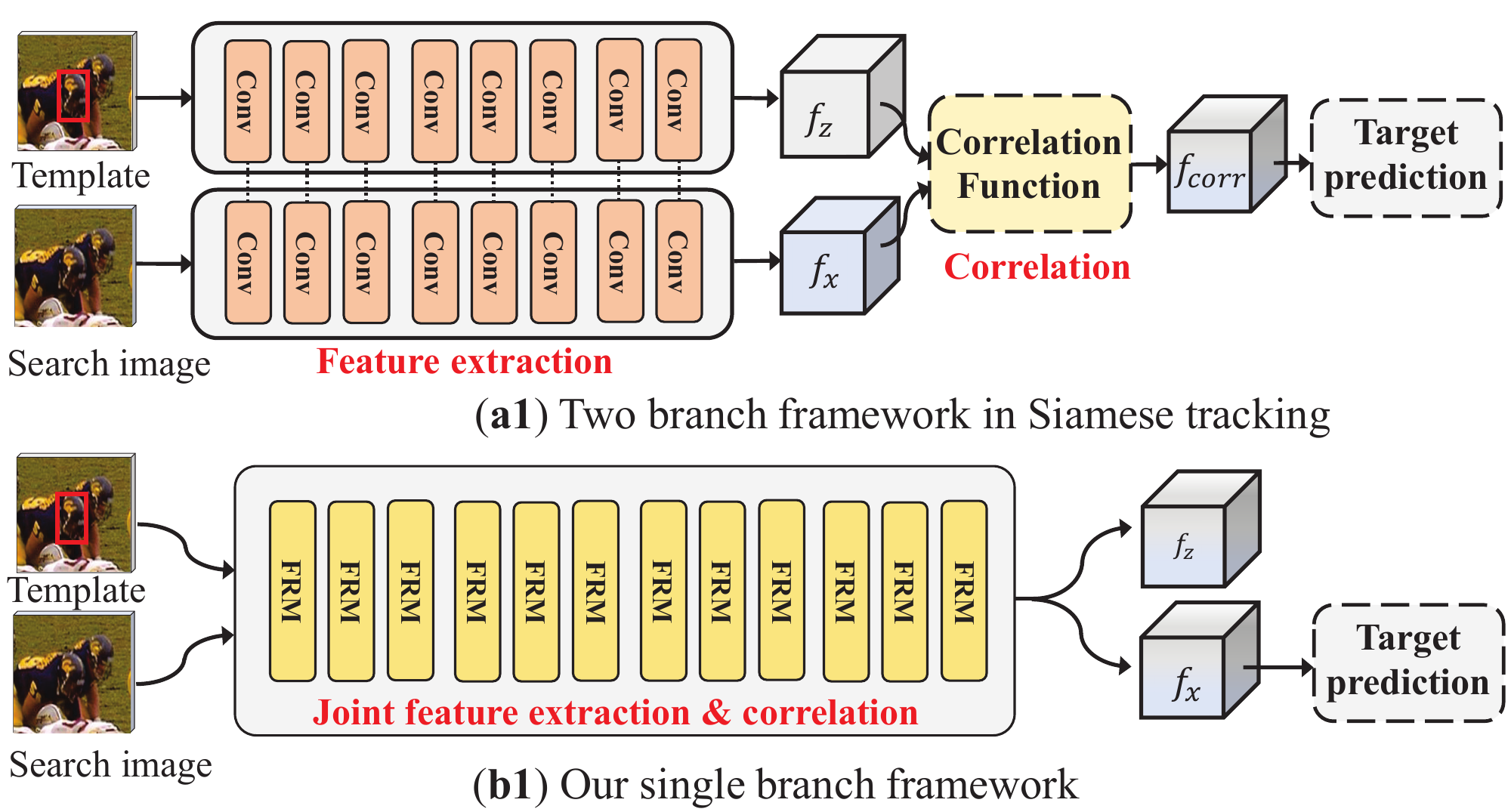} 
\end{minipage}%
\begin{minipage}[t]{0.50\linewidth}
\centering
\includegraphics[scale = 0.43]{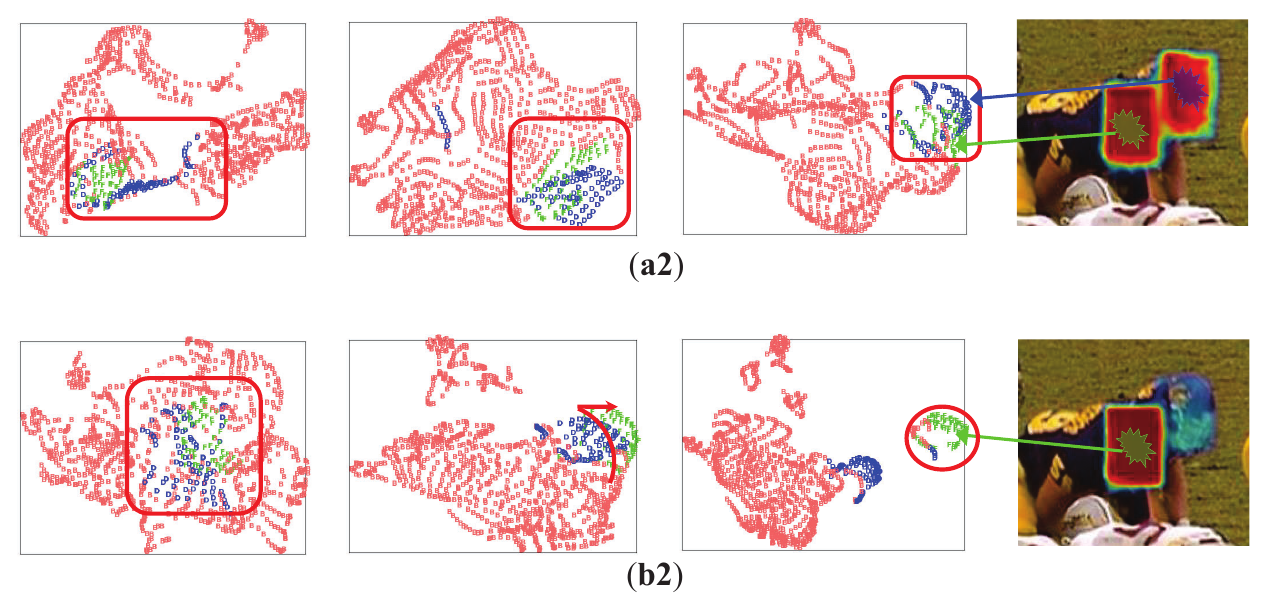} 
\end{minipage}
\caption{
    (a1) Standard Siamese-like feature extraction. (b1) Our single-branch framework uses joint feature extraction and correlation. Our pipeline removes separated correlation steps, e.g., Siamese cropping correlation~\cite{ siamrpn},  DCF~\cite{atom} and transformer-based correlation~\cite{transt};
(a2)/(b2) are the TSNE~\cite{tsne} visualizations of search features in (a1)/(b1) when feature networks go deeper. 
	}
\label{fig:CAcompare}
\end{figure*}

{In the deep learning era, researchers have endeavored to address this challenge from two fundamental perspectives: enhancing the feature network and developing an elaborated correlation operation design.
To achieve this, Siamese-like feature extraction, which utilizes deep convolutional neural networks (CNN) to learn more expressive feature embeddings, has significantly improved tracking performance.
The recent emergence of transformer-based feature networks has further elevated the performance.  
On the other hand, developing a more robust correlation operation can help trackers differentiate targets from similar objects more easily. 
Typical Siamese trackers~\cite{siamfc, siamrpn++} such as SiamRPN~\cite{siamrpn} adopt cross-correlation for efficiency while Discriminative Correlation Filter-based (DCF) trackers~\cite{KCF, DiMP} adopt online filter learning.
 }


{Given that modern CNN-based visual backbones~\cite{alexnet, googlenet, resnet} have become the predominant choice for feature extraction, most trackers are dedicated to developing effective correlation operations that can differentiate targets from distractors based on their features.
Despite their great success, we infer that Siamese-like tracking paradigms are trapped by the target-distractor dilemma caused by the two competing goals mentioned. 
The underlying reasons are as follows:
$1)$ The Siamese encoding process is unaware of the template and search images, leading to weaker instance-level discrimination of learned embeddings.
$2)$ There is no explicit modeling for the backbone to learn the decision boundary that separates the two competing goals, leading to a sub-optimal embedding space.
$3)$ During inference, arbitrary objects, including distractors, can be tracked, whereas each training video only annotates one object.
This gap is further widened by $2)$. 
Our key insight is that feature extraction should have dynamic instance-varying behaviors to generate \textit{appropriate} embeddings for VOT to ease the target-distractor dilemma.}
%

%

%
{
To address these issues, we propose a novel single-branch, fully transformer-based tracking framework that allows for deep interaction between the features of the two images during feature extraction. 
In contrast to the dual-branch Siamese pipeline in Fig.~\ref{fig:CAcompare}(a1), our Single-Branch Transformer tracking (SBT) approach achieves a win-win scenario at the stage of feature extraction, i.e., it distinguishes the target from similar distractors while maintaining coherent characteristics among dissimilar targets.
The effectiveness of the features obtained from SBT has been confirmed through the results presented in Fig.~\ref{fig:CAcompare}(d2).
The overall framework of SBT is shown in~Fig.~\ref{fig:architecture}.
It has four critical architectural components: the Feature Relationship Modeling (FRM) layer, the Patch Embedding (PaE) layer, the Positional Encoding (PE) layer, and the prediction head. 
The template and search images are split into image patches and then projected into feature tokens through the patch embedding layer. 
Feature tokens are fed into the stacked feature relation modeling layer to fuse features within the same image or mix features across images layer by layer. 
FRM is a variant of the transformer layer in vision transformers~\cite{vit, swin, pvt}. 
The search image features obtained from SBT are directly fed to the prediction head. 
Our key technical innovation is the single stream for image pair feature processing that jointly extracts and correlates through homogeneous transformer layers. }

{To better adapt the single-branch pipeline into tracking, we conduct extensive experiments to summarize design principles and propose an improved SBT version, dubbed SuperSBT. 
Specifically, we first introduce a three-stage hierarchical structure to obtain a multi-scale feature representation with a favorable feature downsampling ratio of $16$ in tracking. 
Then, we decouple the layer pattern design in the shallow and deep stages to obtain a more expressive feature embedding for tracking. 
In the deep stage, we design a unified relation modeling layer as the main building unit, which is friendly to cross-image modeling and parallel computing.
The unified relation modeling layer can implement intra-image and cross-image relation modeling by a global attention scheme.
In the shallow stage, we propose a local modeling layer to enhance the local modeling ability, which replaces the attention scheme with convolutional filters. 
We also choose the optimal network variants based on the design principles, balancing speed, performance, and model size.
Three variants of SuperSBT (light, small, and base versions) are proposed and obtain promising results in terms of speed and performance. 
We further explore the temporal modeling ability of SuperSBT via a simple dynamic template scheme, which concatenates an extra template at the beginning of input. 
Moreover, our SuperSBT network is pre-trained via Masked Image Modeling (MIM) on abundant unpaired images such as ImageNet~\cite{ImageNet}, leading to a fast convergence and a better tracking performance in fine-tuning.
In contrast to the SBT baseline, the SuperSBT, with optimal network designs,  obtains superior performance on eight VOT benchmarks, with even faster speed, as shown in Fig.~\ref{fig:bab}. 
}

This study builds upon our conference paper SBT~\cite{sbt} and is one step forward for our conference paper DualTFR~\cite{dualtfr}. It significantly extends them in various aspects. 
For example, we develop an improved tracker dubbed SuperSBT, greatly enhancing the SBT baseline. 
The new contributions are summarized as follows.
\begin{itemize}[leftmargin=0.4cm]

	\item We propose an improved version dubbed SuperSBT, which has more specialized architectural adaptations for visual tracking, which greatly raises the overall tracking performance of our SBT baseline and speeds up online inference.
 
	\item We have proposed more model variants and conducted experiments on more visual tracking benchmarks, e.g., TrackingNet~\cite{trackingnet}, TNL2k~\cite{tnl2k}, and VOT2021~\cite{Kristan_2021_ICCV}. We have also included more comparisons with recent strong trackers.
 
	\item We have provided more thorough design principles, discussions, and details.

\end{itemize}

\section{Related Work}
\label{sec:related}

To provide context for our work, we discuss some of the most representative developments in Siamese-based tracking, vision transformers of the last few years, and the application of transformers in visual tracking.

\subsection{Siamese Tracking}
Prior to the prevalence of Siamese tracking, Discriminative Correlation Filter (DCF) trackers~\cite{KCF, ECO, BACF} perform visual tracking by learning a target model online. 
Though online solving least-squares based regression is further improved by fast gradient algorithm~\cite{atom}, end-to-end learning~\cite{DiMP, dcfst} and CNN-based size estimation~\cite{atom, DiMP, tomp}, 
DCF methods remain highly sensitive to complex handcrafted optimization, impeding their ability to achieve.

In recent years, Siamese-based methods~\cite{siamfc, siamrpn, siamrpn++, siamdw, siamfcpp,SiamCAR, siamban} attract great attention in the tracking field due to a more optimal trade-off between accuracy and efficiency. 
The pioneering work of SiamFC~\cite{siamfc} constructs a fully convolutional Siamese network to formulate visual tracking as pair-wise template matching. 
After that, SiamRPN~\cite{siamrpn} combines the Siamese network with RPN~\cite{fast_rcnn, fasterrcnn} and conducts template matching using a simple depthwise~\cite{siamrpn++} correlation to obtain more precise tracking results. 
Under the Siamese-based framework, Siamese trackers are improved with the help of the following techniques: powerful backbones~\cite{siamrpn++, siamdw}, elaborated prediction networks~\cite{siamrpn, siamfcpp, SiamCAR}, attention mechanism~\cite{attnsiam, stmtrack} and model fine-tuning~\cite{ targetaware, targetaware1, MAML}, cascaded frameworks~\cite{spm, CascadedSiameseRPN, siamrn}.

Despite the improvements, most Siamese trackers~\cite{siamrpn, siamrpn++, siamdw, siamfcpp,SiamCAR, siamban} still follow a similar pipeline: a backbone network to extract image features and a correlation step to compute the similarity between the template and the search region. 
However, few of these Siamese tracking methods 
notice that independent feature extraction and correlation pipeline put the trackers into a target-distractor dilemma.
The Siamese encoding process is unaware of the template and search images, which weakens the discrimination of learned embeddings.
Furthermore, there is no explicit modeling for the backbone to learn the decision boundary that separates the two competing goals.
Instead, our target-dependent feature network can greatly ease the dilemma by unifying the feature extraction and correlation step into a single-branch transformer model. 
SBT feature network can explicitly model the relationship between the template and search images, thus formulating a novel and conceptually simple tracking pipeline by removing the separated correlation step in Siamese trackers.

\subsection{Vision Transformer} 
CNNs~\cite{alexnet, googlenet, resnet} generally serve as the standard feature extraction network throughout computer vision.
Recently, inspired by the success of self-attention~\cite{attn} layers and transformer~\cite{detr} architectures in the NLP field, some works employ self-attention layers to replace some or all of the spatial convolution layers in the CNN structure, which formulate a hybrid CNN-Transformer architecture~\cite{detr, transpose, transt}.
Then, Vision Transformer (ViT)~\cite{vit, swin, pvt}, constructed by a pure-transformer model, achieves impressive results as a vision backbone.
Deeper and more effective architectures are the two pillars of powerful backbones, which boost numerous downstream tasks. 
Similarly, the improvements brought by the powerful backbone in VOT are mainly attributed to the more expressive feature embedding~\cite{siamdw, siamrpn++}, which has subtle differences from other tasks, e.g., object detection. 
However, the dynamic nature of VOT requires asymmetrical encoding for template and search images, which has not been given sufficient attention in most prior works.  
Considering that, we propose a dynamic instance-varying backbone for VOT beyond only pursuing an expressive embedding.

\begin{figure*}[t]
	\centering{\includegraphics[scale = 0.53]{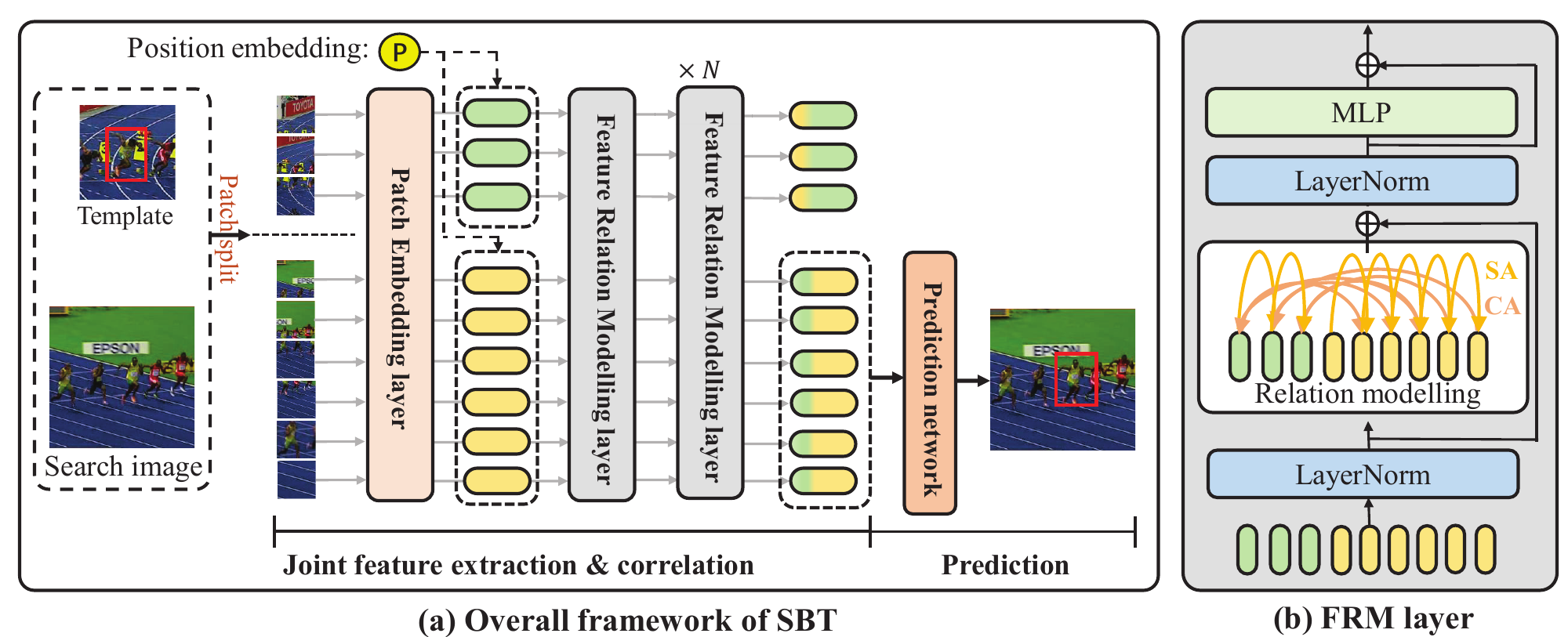}}
	\caption{(a) Architecture of our proposed Single-Branch Transformer framework for tracking (SBT). 
 Unlike Siamese, DCF, and Transformer-based methods, it has no standalone module for computing correlation. Instead, it embeds correlation in all Feature Relation Modeling (FRM) layers at different network levels. 
The fully fused features of the search image are directly fed to the prediction network to obtain the localization and size of the target. 
(b) shows the structure of the FRM layer, which is a variant of the transformer~\cite{vit} layer.
There are two options for attention operators in the FRM layer, i.e., Self-Attention (SA) and Cross-Attention (CA).
SA  operator fuses features within the same image while the CA operator mixes features across images.
}
	\label{fig:architecture}
	\vspace{-4mm}
\end{figure*}

\subsection{Transformer in Tracking}
Several efforts have been made to apply the transformer model to the tracking field. 
The pioneer works follow a hybrid CNN-transformer architecture, which enhances original CNN-based trackers with transformer modules. 
TransT~\cite{transt} replaces a shallow correlation step with a transformer network, fusing the template 
and search features extracted by a CNN backbone.
TMT~\cite{tmt} employs the transformer and combines it with SiameseRPN~\cite{siamrpn} and DiMP~\cite{DiMP} as a feature enhancement module rather than replace the correlation. 
Stark~\cite{stark} also follows a hybrid CNN-Transformer architecture, which fuses the information by concatenating the search region and the template. 
Then, DualTFR~\cite{dualtfr} and SwinTrack~\cite{swintrack} replace the original CNN-based backbone with vision transformers~\cite{vit, swin}, which construct pure transformer trackers. 
However, direct replacement of the backbone cannot fully exploit the power of the transformer model. 
Instead of using a transformer as a fusion module~\cite{transt, stark, dtt} or backbone, we leverage the vision transformer to extract and correlate the template and search features jointly. 
Our single-branch transformer-based tracker greatly simplifies the hybrid CNN-Transformer and leverages the attention scheme more thoroughly for visual trackers.

The concurrent works OStrack~\cite{ostrack}, Mixformer~\cite{mixformer}, and SimTrack~\cite{simtrack} also share a similar idea with our SBT tracking. 
The difference is that our work conducts extensive studies on Vision Transformers for tracking and developing the optimal model variants for visual tracking. 
In contrast to directly applying Vision Transformer to tracking, i.e., OStrack~\cite{ostrack} and Mixformer~\cite{mixformer}, 
we modify the key modules in Vision Transformer, e.g., attention scheme and hierarchical structure, to better adapt the transformer to tracking. 
Our improved fully transformer-based tracker, dubbed SuperSBT, achieves better tracking performance while running at high speed.

\section{Single-Branch Transformer Tracking}
\label{sec:Architecture}
In this section, we first illustrate the overall architecture of our Single-Branch Transformer (SBT) tracking as shown in Fig.~\ref{fig:architecture}.
Then, we present the details of two fundamental building components: a transformer-based backbone and a prediction network. 
Finally, we give exploration studies on finding optimal SBT variants and summarize effective principles.
%
%

\subsection{Overall Architecture of SBT}
In contrast to previous Siamese tracking, SBT tracking simplifies the tracking pipeline by leveraging the transformer-based backbone for joint feature extraction and fusion. 
SBT firstly splits an image pair, comprising a template image $z \in \mathbb{R}^{3 \times H_{z} \times W_{z}}$ and a candidate search image $x \in \mathbb{R}^{3 \times H_{x} \times W_{x}}$, into non-overlapping image patches $\{N_{x}, N_{z}\}$ through the Patch Embedding layer (PaE) $\mathbf{E}$.
In general, $z$ is centered on the target object while $x$ represents a larger region in the subsequent frame that contains the target.
The feature tokens embedded from image patches $\{N_{x}, N_{z}\}$, construct a sequence of token as the input:
\begin{equation}
f_{zx} =[f_{z}, f_{x}]= [\mathbf{E}x_1, \ldots , \mathbf{E}x_{N_{x}}, \mathbf{E}z_1, \ldots , \mathbf{E}z_{N_{z}}] + \mathbf{PE},
\label{eq:vit_embed}
\end{equation}
where $\mathbf{PE}$ is a positional embedding to differentiate each token. 
%
The tokens are passed through a transformer backbone with a hierarchical/plain structure of $L$ relation modeling layers.
Each layer $\ell$ comprises of Multi-head Self-Attention (MSA)~\cite{attn}, Layer Normalisation (LN)~\cite{LN}, and a Multi-Layer Perceptron (MLP)~\cite{vit} as follows:
\begin{equation}\label{eq:vittrack}
\begin{aligned}
y_{zx}^{\ell} &= \text{MSA}(\text{LN}(f_{zx}^\ell)) +f_{zx}^\ell,  \\
f_{zx}^{\ell + 1} &= \text{MLP}(\text{LN}(y_{zx}^\ell)) + y_{zx}^\ell,
\end{aligned}
\end{equation}
where the MLP consists of two linear projections separated by a GELU~\cite{gelu} non-linearity. 
Then, the template and search image features $\{f_{z}, f_{x}\}$ are jointly extracted and fused through multiple relation modeling layers. 
In the final stage, a prediction network is used to decode the fused search image feature $f_{x}^L$ to locate and estimate the size of the target:
\begin{equation}
\begin{aligned}
& y_{reg} = \Phi_{reg}(f_{x}^L), \quad y_{cls} =\Phi_{cls}(f_{x}^L), 
\end{aligned}
\end{equation}
where $\{y_{reg}, y_{cls}\}$ denote the target location and shape estimation results. $\{\Phi_{cls}, \Phi_{reg}\}$ are classification and regression head.

%
\emph{Differences with CNN-based Trackers.} 
SBT introduces fewer paddings than CNN-based trackers.
Deep trackers have an intrinsic requirement for strict translation invariance, 
$f\left(c, x\left[\Delta \tau_{j}\right]\right)=f(c, x)\left[\Delta \tau_{j}\right]$
where $\Delta \tau_{j}$ is the shift window operator, $c$ denotes the template/online filter in Siamese/DCF tracking, and $f$ denotes correlation operation. 
Modern backbones~\cite{resnet, resnext, MobileNets} can provide more expressive features while 
their padding does not maintain translation invariance.
Thus, deep trackers~\cite{siamdw, siamrpn++} crop out padding-affected features and adopt a spatial-aware sampling training strategy to keep the translation invariance.
Theoretically, padding in SBT can be removed completely or only exists in patch embedding for easy implementation. 
Moreover, the flattened feature tokens have permutation invariance, making the SBT completely translation invariant. 
%
%
Thus, we argue that SBT-driven tracking can theoretically overcome the intrinsic restrictions in classical deep trackers by using brand-new network modules. 

\subsection{Transformer-Based Backbone}
Our SBT is on top of a transformer-based backbone with three key components: patch embedding layer, feature relation modeling layer, and positional encoding methods. 

\smallskip\noindent\textbf{Patch embedding.}
\label{sec:Embedding}
%
The patch embedding layer is to embed the two images into feature tokens through a convolutional layer.
The kernel size and stride of the convolutional layer in the patch embedding layer can be modified for different model structures, and we conduct exploration studies in Sec.~\ref{sec:empirical}. 
%
%

%

\begin{table*}[t]
    \begin{center}
        	\resizebox{2.0\columnwidth}{!}{%
        \fontsize{8.0 pt}{3.5mm}\selectfont
        \begin{threeparttable}
                \setlength{\tabcolsep}{0.01mm}

            \begin{tabular}{ @{}c@{}  
            @{}c@{} @{}c@{} @{}c@{} @{} c@{} 
            @{}c@{} @{}c@{} @{}c@{} | @{}c@{} 
            @{}c@{}  @{}c@{} @{}c@{} @{}c@{} 
            @{}c@{} @{}c@{} @{}c@{} @{}c@{} 
            @{}c@{} @{}c@{} @{}c@{} @{}c@{}}
                \toprule[0.08em]
                \textbf{Setting}
                & ~$\rm A_{1}$\tnote{1}~ & ~$\rm A_{2}$\tnote{2}~ & ~$\rm A_{3}$~
                & ~$\rm A_{4}$~ & ~$\rm A_{5}$~  & ~$\rm A_{6}$~
                & ~$\rm A_{7}$~
                & ~~  
                & \textbf{Setting}
                & $\rm B_{1}$ & $\rm B_{2}$ & $\rm B_{3}$  
                & $\rm B_{4}$ & $\rm B_{5}$ &$\rm B_{6}$ 
                & $\rm B_{7}$ & $\rm B_{8}$
                \\
                \midrule[0.06em]
               \textbf{Refer to}
                & ViT~\cite{vit}~
                & ~~SwinT~\cite{swin}~
                & ~~ResT~\cite{rest}~ 
                & ~~PVT~\cite{pvt}~ 
                & ~~PVT~\cite{pvt}~ 
                & ~~PVT~\cite{pvt}~ 
                & ~~Twins~\cite{twins}~
                & ~~  
                &~\textbf{DIM(1,2)}~
                &[64, 128] &[64, 128] &[64, 128]
                & [64, 128] & [64, 128] & [64, 128]
                & [64, 128]  
                & [32, 64]

                \\
                ~\textbf{ATTN}~
                &~VG~ &~SL~ & ~SRG~ &~SRG~
                &~SRG~ &~SRG~ &~VL/SRG~

                & ~~  
                
                &~\textbf{DIM(3,4)}~
                &~[320]~ & ~[320,512]~ & ~[320,512]~
                & ~[512]~ &~[320]~ &~[320,512]~
                &~[320]~
                 &~[320]

                \\
                \textbf{PE} &~Abs
                & ~Rel~ & ~Cond~
                & ~Cond~ & ~Cond~
                & ~Rel~ & ~Cond~
                & ~~  
                &\textbf{BLK}
                &[3,4,10] & [4,2,6,1] & [2,2,6,2]
                & [2,2,4] & [3,4,10] & [2,4,6,1]
                & [3,4,12] 
                & [3,4,10] 
                \\
                \textbf{PaE}
                & ~$P_{1}$\tnote{3}~ & ~$P_{2}$\tnote{3}~ & ~~Conv~ & ~$P_{2}$\tnote{3}~ 
                & ~~Conv~ & ~~Conv~ & ~~Conv~
                & ~~  
                &\textbf{STR}
                & [4,2,1] & [4,2,1,1] & [4,2,1,1] 
                & [4,2,1] & [4,1,2] & [4,2,1,1] 
                & [4,2,2]\tnote{4}
                & [4,2,1] 
                \\
                
               \midrule[0.06em]

                ~~\textbf{Param.(M)}~ 
                & ~~21.5~ 
                & ~40.2~ & ~23.9~
                &  ~20.1~ & ~21.3~ &  ~21.0~
                & ~19.6~
                &~~  
                &~~\textbf{Param.(M)}~ 
                &~~21.3~& ~18.6~ &\cellcolor{red!5}~21.1~
                & ~20.5~&~20.8~ & ~19.3~
                & ~20.8~
                 & ~15.1~
                \\
                ~\textbf{Flops(G)}~
                & ~~10.6~ & ~36.5~ & ~20.2~ 
                & ~18.9~ & ~19.6~ &  ~19.3~
                &  ~17.5~
                & ~~  
                &~\textbf{Flops(G)}~
                & ~~19.6~ & ~19.3~ & ~22.5~
                &  ~19.2~ & ~24.4~ & ~24.7~
                &  ~12.1~
                 & ~14.5~
                \\
                 \textbf{AO}
                & ~~~58.2~ & ~62.4~ & ~63.7~
                & ~61.7~ &~63.5~ &~63.1~
                & ~60.1~ 
                & ~~    
                & \textbf{AO} 
                &  ~~~63.5~ & ~57.4~ &~60.9~ 
                & ~56.7~ & ~63.3~ & ~60.6~  
                & ~52.2~
                &~56.2~
                \\
                \bottomrule[0.08em]
                
            \end{tabular}

            \begin{tablenotes}
                \footnotesize
                \item[1] $\rm A_{1}$ does not have a hierarchical structure, so we adopt $16$ downsampling ratio at the beginning and reduce the number of transformer layers to have a comparable model size.
                \item[2] For $\rm A_{2}$, we set the same image size $(224 \times 224)$ for template and search image for simplicity.
                \item[3] $P_{1}$ denotes the $\rm A_{1}$ splits an input image into non-overlapping patches ($4 \times 4$) and changes feature dimension with a linear layer. $P_{2}$ denotes patch merging, which changes feature dimension after patch split. Conv denotes the strided convolutional layer.  
                \item[4] For model settings with total network stride 16, we increase the search image size to $320 \times 320$ for a fair comparison.   
            \end{tablenotes}
\caption{The left part compares different factors of SBT, including attention computation methods (ATTN), position encoding methods (PE), patch embedding methods (PaE), number of model parameters, and flops. The right part compares the rest of the factors based on $\rm A_5$ (described in the left part), such as the feature dimensions (DIM) and the number of blocks (BLK), as well as the stride of the feature maps in each stage. All models, unless explained, follow the same setting: training from scratch, interleaved FRM-SA/FRM-CA block in the third stage, $128$ for template image, and $256$ search image. All the experiments follow the official GOT-10k~\cite{got} test protocol.}
\label{tab:vit}
        \end{threeparttable}
        
        }
	\vspace{-2mm}
    \end{center}

\end{table*}

\smallskip\noindent\textbf{Feature relation modeling.} 
\label{sec:CASA}
%
The Feature Relation Modeling (FRM) layer can simultaneously model cross-image and intra-image relationships.
As shown in Fig~\ref{fig:architecture}(b), the FRM layer follows the structure of a standard transformer layer~\cite{attn, vit}. 
Intuitively, the FRM layer gradually fuses features from the same and different images through Self-Attention (SA) and Cross-Attention~(CA), respectively. 
In each FRM layer, the template  $f_{z}$ and search features $f_{x}$  are reshaped to feature tokens and then embedded into the Query/Key/Value space.
Let $\chi_{(.)}$ denote a function that reshapes/arranges feature maps into the desired form. The function varies for different methods. We compute the $q, k, v$ features as:
\begin{equation}
\begin{aligned}
& q_{i}= [\chi_{q} (f_{i})] ^\mathsf{T}\omega_{q},\quad i \in \{z, x\}, \\
& k_{i}= [\chi_{k} (f_{i})] ^\mathsf{T}\omega_{k},\quad i \in \{z, x\}, \\
& v_{i}= [\chi_{v} (f_{i})] ^\mathsf{T}\omega_{v},\quad i \in \{z, x\}, 
\end{aligned}
\end{equation}
where $\{\omega_{q},\omega_{k},\omega_{v}\}$ represent linear projections. 
Many works~\cite{pvt, swin} attempt to modify the attention operators with task priors. 
%
For example, the Vanilla Global attention (VG)~\cite{vit} computes attention among all feature tokens. 
So $\{\chi_{q},\chi_{k},\chi_{v}\}$ represent identity mapping. 
The Spatial-Reduction Global attention (SRG)~\cite{pvt, rest} uses a convolution with a stride larger than one (i.e., $\{\chi_{k},\chi_{v}\}$) to reduce the spatial resolution of the key and value features. 
The resolution of the query features has not been changed. Then, it computes global attention as VG. 
The Vanilla Local window attention (VL)~\cite{twins} split feature tokens in groups based on their spatial locations and only compute attention within each group.
Swin Transformer~\cite{swin} further adds a Shift window mechanism to vanilla Local attention~(SL) for global modeling. 
More empirical studies and discussions are in Sec.~\ref{sec:empirical}. 
Here, we omit the various attention operators in the formula for clarity. 
The following equation shows how we compute self-attention and cross-attention to model intra-image and cross-image relationships: 
\begin{equation}
\begin{aligned}
& \tilde f_{ij}  = {\rm Softmax} (\frac{q_{i} k_{j}^\mathsf{T} }{\sqrt{d_{h}}}) v_{j}, \quad i, j \in \{z, x\}, \\
\end{aligned}
\end{equation}
In SA, $i$ and $j$ are from the same source (either $z$ or $x$) and the resulting feature update is:
\begin{equation}
\begin{aligned}
& f_{z} :=  f_{z} + \tilde f_{zz},\quad f_{x} :=  f_{x} + \tilde f_{xx}, 
\end{aligned}
\end{equation}
In CA, it mixes the features from different sources: 
\begin{equation}
\begin{aligned}
& f_{z} :=  f_{z} + \tilde f_{zx},\quad f_{x} :=  f_{x} + \tilde f_{xz}.
\end{aligned}
\end{equation}

We can see that the correlation between the two images is deeply embedded in feature extraction seamlessly.
Thus, SBT tracking can jointly extract and correlate template and search features without an extra correlation step. 

\emph{Differences with convolutional-based correlation.}
Cross-attention conducts feature interaction more than twice.
We first prove that Cross-attention can be decomposed into dynamic convolutions.
Cross-attention, which performs as feature correlation, is mathematically equivalent to two dynamic convolutions and a SoftMax layer. 
For simplicity, we annotate the encoded $\{q,k,v\}$ features to their original feature as the projection matrix is $1 \times 1$ position-wise convolutional filters. 
So the cross-attention for query from search feature $x$ to template feature $z$ is:
\begin{equation}
\begin{aligned}
&{\rm x_z} = {{\text{RS}} (z)}^Tx+\bm{0} = {\bm W_{1}}(z)^T x+\bm{b_{1}}, \\
&{\text{Attn}}_{xz} = {\text{SoftMax}} ({\rm x_z}),\\
 &\tilde f_{xz} = {{\text{Attn}_{xz}} z} + x = \bm{W_{2}}(z, x)^Tx+{\bm b_{2}}(x),
\end{aligned}
\label{early_attn}
\end{equation} 
where $\bm{W}(a, b), \bm{b}(a, b)$ is the weight matrix and bias vector of dynamic filters generated by $\{a, b\}$ and $\rm RS$ denotes reshape. 
To obtain the correlation feature $\tilde f_{xz}$, the search feature $x$ goes through a dynamic convolutional layer generated by $z$, a SoftMax layer, and another dynamic convolutional layer generated by $z$ and $x$. Two dynamic convolutional layers come from reshaping $\rm z$ along the channel and spatial dimension. 
The depth-wise correlation or pixel-wise correlation~\cite{AlphaRefine} is only equivalent to one dynamic convolutional layer.
Thus, cross-attention is twice as effective as the previous correlation operator with the same template feature as dynamic parameters.

\smallskip\noindent\textbf{Position encoding.}
Since template and search features are reshaped into feature tokens,  Positional Encoding (PE) embeds spatial information into sequential tokens. 
Non-parametric or conditional methods can implement positional encoding.
For majority methods~\cite{vit, swin, detr}, the encoding is generated by the sinusoidal functions with Absolute coordinates~(Abs) or Relative distances~(Rel) between tokens.
Being much simpler, Conditional positional encoding~\cite{condPE, pvt, rest}~(Cond) generates dynamic encoding by convolutional layers.
In our SBT model, we add a $3 \times 3$ depth-wise convolutional layer $\varphi_{pe}$ to MLP before GELU as conditional positional encoding.
We further improve it by using relative positional encoding.
More details are presented in Sec.~\ref{sec:empirical}.

\begin{figure*}[t]
	\centering{\includegraphics[scale = 0.51]{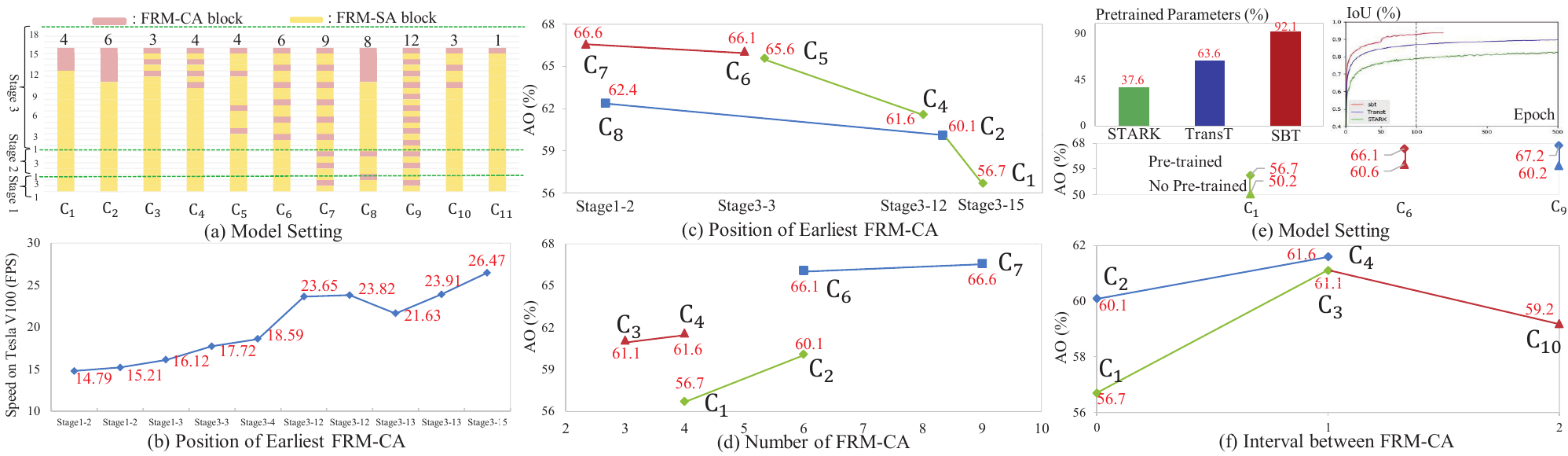}}
	\caption{Studies on the number/position of FRM-CA block. (a) Different model settings, (b) Speed vs. different model settings,(c) Tracking performance vs. position of earliest FRM-CA layer, (d) Tracking performance vs. number of FRM-CA layers, (e) Tracking performance vs. pre-trained or not, (f) Tracking performance vs. intervals between the FRM-CA layer.}
	\vspace{-2mm}
	\label{fig:main}
\end{figure*}

\subsection{Direct Prediction Head}
Differing from existing tracking methods, we directly attach a classification head and regression head to the search feature from the SBT feature network without any additional correlation operations.
%
In this work, we adopt a lightweight convolutional-based anchor-free head~\cite{centernet, SiamCAR, ostrack}, as illustrated in Sec.~\ref{sec:experiment}.

\emph{Differences with feature pyramid-based prediction head.}
The serial pipeline method allows the prediction network to utilize hierarchical features.
Siamese trackers~\cite{siamrpn++, siamban} correlate each hand-selected feature pair and feed them into parallel prediction heads. Then, prediction results are aggregated by a weighted sum.    
Compared to the handcraft layer-wise aggregation, the SBT structure intrinsically explores multi-level feature correlation. We take three-level feature utilization as an example: 
\begin{equation}
\begin{aligned}
{x}_{i}, {z}_{i}&= \text{FRM}_{\text{CA}}^{i}({\tilde {x}_{i}}, {\tilde {z}_{i}}), \quad i \in \{0, 1, 2\}\\
{x}_{2}, {z}_{2} &= \text{FRM}_{\text{CA}}^{2} ( \text{FRM}_{\text{CA}}^{1} ( \text{FRM}_{\text{CA}}^{0} ( {x}_{0}, {z}_{0} ))),\\
{x}_{p}&= \Phi_{p}({{x}_{2}}),
\end{aligned}
\label{early}
\end{equation} 
where $\{0, 1, 2\}$ represents shallow, intermediate and deep level, $\{\tilde {x}, \tilde {z}\}$ are the previous layer features of $\{{x},  {z}\}$, $\{\text{FRM}_{\text{CA}}, \Phi_{p}\}$ denote FRM-CA block and prediction head. 
Using a serial pipeline, the predicted result ${x}_{p}$ inherently contains hierarchical feature correlations.

\begin{figure*}[t]
	\centering{\includegraphics[scale = 0.53]{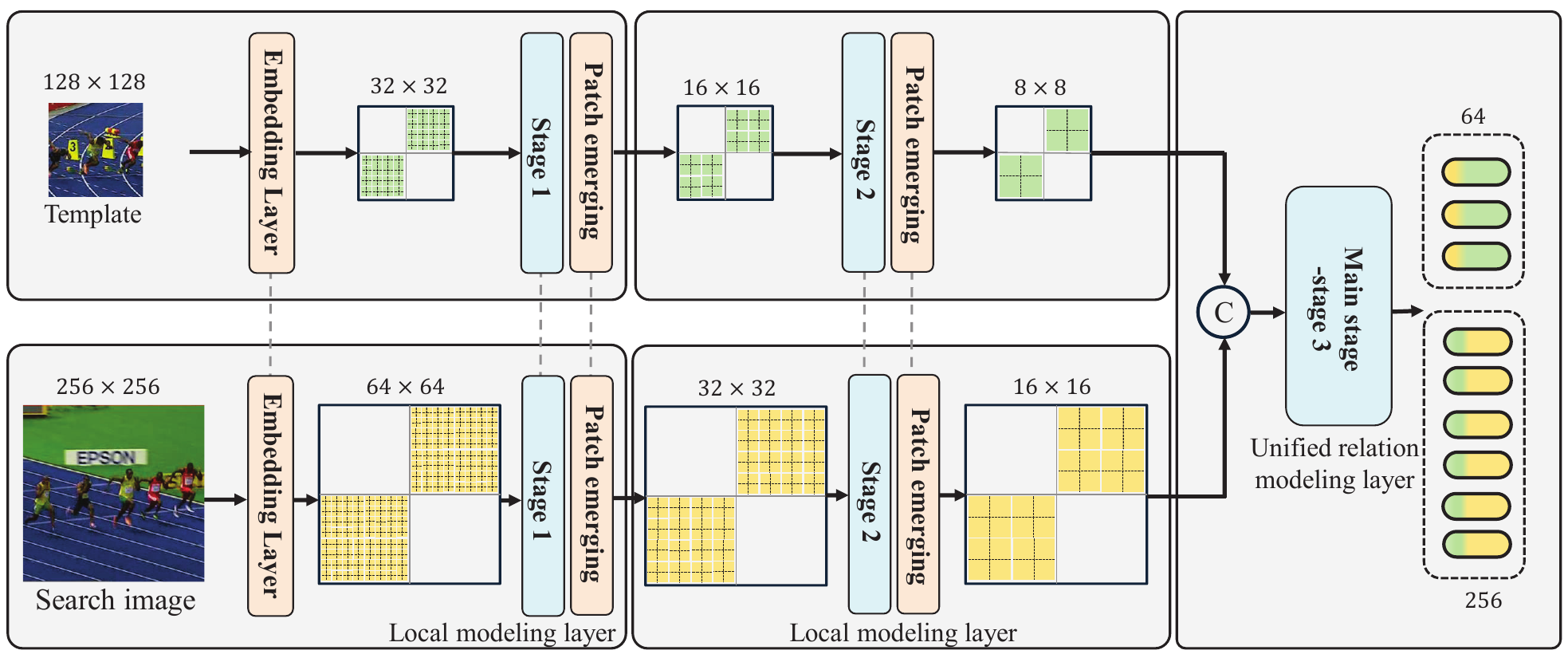}} 
	\caption{Architecture of our improved Single Branch Transformer framework for tracking (SuperSBT). Based on the summarized design principles, we upgrade the SBT baseline with a local modeling layer, unified relation modeling, and reasonable architecture variants. 
}
	\label{fig:supersbt_arch}
	\vspace{-4mm}
\end{figure*}

\subsection{Exploration Studies on SBT}
\label{sec:empirical}
In this section, we analyze the impact of different structure designs in SBT tracking and assess their performance. The effective design principles are summarized below, which can be used to enhance the SBT tracking pipeline in Sec.~\ref{sec:supersbt}.
%

\paragraph{Principle A.} 
\emph{Hierarchical transformer structure is more effective than plain structure.} 
Based on the findings presented in Tab.~\ref{tab:vit}, it is evident that the model variants from $\rm A_{2}$ to $\rm A_{7}$ outperform the single-stage model variant $A_{1}$, despite owning a similar model size. This suggests that a hierarchical structure is more effective due to its ability to provide a multi-scale representation.

\paragraph{Principle B.} \emph{The attention scheme should be friendly to both cross-image modeling and parallel computing.} 
The primary distinction in attention computation is the operation to reduce complexity~(global/local attention). 
We find that the local attention (VL/SL) block cannot perform Cross-Attention directly due to the inequality of local windows in the template and search image. This harms high-speed parallel computing.
Therefore, we use the same image size of $224 \times 224$ for both template/search images ($\rm A_{2}$) for SBT constructed by pure local attention layers to avoid complex cross strategies.
Comparing to the settings with global attention block (VG/SRG) ($\rm A_{3}$ to $\rm A_{7}$), which adopt $128\times 128$ as template size, the performance of pure local attention ($\rm A_{2}$) drop at least $3.6$ points in $AO$ with more parameters and flops. 
This is mainly because the template contains too much background information, which may confuse the search algorithm. 
%


\paragraph{Principle C.} \emph{Early-cross helps the tracker to see better.} 
Based on the aforementioned principles, SBT could reap the advantages of increased cross-correlation at an earlier model stage.
We remove different positions/numbers of the FRM-CA layer in Fig.~\ref{fig:main}.
As shown in Fig.~\ref{fig:main}(d), when the number of FRM-CA layers increases, the model's performance consistently improves with the same FRM-SA/FRM-CA position pattern ($\rm C_{3}$ vs. $\rm C_{4}$, $\rm C_{1}$ vs. $\rm C_{2}$, $\rm C_{6}$ vs. $\rm C_{9}$). 
It proves that the SBT tracker benefits from a more comprehensive Cross-Attention between template and search branches.
In Fig.~\ref{fig:main}(d), when the number of FRM-CA layers is the same, earlier cross design has significant positive impacts ($\rm C_{4}$ surpasses $\rm C_{1}$ by $4.9$ points, $\rm C_{6}$ surpasses $\rm C_{2}$ by $6.5$ points).
The underlying reason is that early-cross generates better target-dependent features.
Benefitting from the full transformer structure, SBT converges much faster than ``CNN+transformer" trackers~\cite{transt, stark}, as shown in Fig.~\ref{fig:main}(e).

\paragraph{Principle D.} \emph{Layer patterns in the shallow and deep stages can be designed differently.} 
The optimal placement pattern significantly impacts the performance of FRM-CA layers.
So, we attempt to place the FRM-SA/FRM-CA layer differently. 
In Fig.~\ref{fig:main}(f), we surprisingly find that the interleaved FRM-SA/FRM-CA pipeline performs better than the separation pattern even with less Cross-Attention and latter earliest cross position ($\rm C_{3}$ vs. $\rm C_{1}$).
The potential cause is that the FRM-SA layer can refine the template/search feature after the correlation, resulting in a more expressive feature space for matching. 
In Fig.~\ref{fig:main}(f), model ($\rm C_{9}$) achieves the best performance $67.2\%$ when the interval is $1$. 
When the interval increases to 2, the performance drops from  $61.1\%$ to $59.2\%$ ($\rm C_{3}$ vs. $\rm C_{10}$).
%
%
From Fig.~\ref{fig:main}(b) and Fig.~\ref{fig:main}(c), we observe that early-cross in shallow-level (stage 1 and 2) does not bring many improvements ($\rm C_{2}$ vs. $\rm C_{8}$, $\rm C_{6}$ vs. $\rm C_{7}$) but lowers the inference speed. 
This is because initiating the cross too early disrupts the one-shot inference, and the shallow feature is less expressive.
%

\paragraph{Principle E.} \emph{Network variants significantly influence speed, performance, and model size.} 
%
%
We must choose the optimal network stride, model stage, and size to design an effective deep tracker.
As shown in Tab.~\ref{tab:vit}, overparameters and flops in shallow stage levels are harmful.
It is mainly because the low dimension can not formulate informative representations ($57.4$ of $\rm B_{2}$ vs. $60.6$ of $\rm B_{6}$). 
We observed a slight improvement in performance with increasing head number, but it decreased speed.
With the same total network stride, the three-stage model performs better than the four-stage model ($63.5$ of $\rm B_{1}$ vs. $57.4$ of $\rm B_{2}$) with comparable parameters and flops. 
Though setting the network stride to 16 can reduce the flops, the performance drops $11.3$ points ($\rm B_{1}$ vs. $\rm B_{7}$), indicating that SBT tracker prefers larger spatial size of features.
It is crucial to achieving a balance between the number of blocks and channel dimensions as they significantly impact the model's size ($56.7$ of $\rm B_{4}$ vs. $63.3$ of $\rm B_{5}$).

\paragraph{Principle F.} \emph{Position encoding and patch embedding layer can be flexibly designed for SBT tracking.} 
We test three positional encoding methods, i.e., absolute, relative, and conditional positional encoding, finding that the difference among them is relatively small.
For instance, Conditional PE only surpasses the relative PE by $0.4$ points ($\rm A_{5}$ vs. $\rm A_{6}$) while Conditional PE brings extra model parameters.  
Thus, we conclude that PE does not significantly impact performance and can be designed flexibly.   
Then, we ablate three cases of patch embedding: patch merging layer and convolutional layers with small or large strides.
We find that the convolutional layer with a small stride is more expressive than the patch merging layer~($\rm A_{4}$ vs. $\rm A_{5}$).


\section{Improved Single-Branch Tracking}
\label{sec:supersbt}
In this section, we first provide an overview of improvements in SuperSBT. 
Then, we present the architectural details of SuperSBT.  
Finally, we introduce network variants of SuperSBT in Tab.~\ref{tab:sbt_variant}.

\subsection{Overview of Improvements}
The overall architecture of SuperSBT is illustrated in Fig.~\ref{fig:supersbt_arch}. 
Following \textbf{Principle A} and \textbf{Principle D}, SuperSBT adopts a hierarchical transformer architecture (see Sec.~\ref{sec:hier}) and independent layer design for shallow and deep model stages. 
Specifically, we stack local modeling layers (see Sec.~\ref{sec:lml}) and patch merging layers in the first two stages, gradually reducing the number of tokens and expanding the channel dimension.  
It is worth noting that the initial two-stage processes involving the template and search image tokens are separated.
Based on \textbf{Principle B}, we propose a unified relation modeling layer (see Sec.~\ref{sec:grm}) for joint feature extraction and correlation in the third stage. 
Following \textbf{Principle~C} and \textbf{Principle E}, we select the optimal architecture variants (see Sec.~\ref{sec:variant}), such as layer number and channel dimension, to improve both performance and speed.
Additionally, we chose patch-merging and a convolutional layer as our PaE method based on \textbf{Principle F}. In the final, SuperSBT is further enhanced by masked image modeling pre-training (see Sec.~\ref{sec:mim}), a simple temporal modeling scheme (see Sec.~\ref{sec:temporal}), and an enhanced prediction head (see Sec.~\ref{sec:mixmlp}). 

\subsection{ Hierarchical Architecture}
\label{sec:hier}
The SuperSBT network can be divided into three stages to produce a hierarchical representation.
The image pairs are first embedded into tokens by a convolutional layer.
Then, patch-merging layers reduce the number of tokens as the network gets deeper. 
Between every two stages, the patch merging layer concatenates the features of each group (dimension $C$) of $2\times 2$ neighboring patches and applies a linear layer on the $4C$-dimensional concatenated features.
This reduces the number of tokens by a multiple of $2\times2=4$ ($2\times$ downsampling of resolution), and the output dimension is set to $C$. 
In this way, the scales of the three stages become 
$\frac{1}{4}$, $\frac{1}{8}$ and $\frac{1}{16}$ of original scale, respectively.
%

%
In contrast to popular hierarchical architecture design in Swin Transformer~\cite{swin} and PVT~\cite{pvt}, we modify the architecture in the following aspects to better adapt it to tracking tasks:
1) We adopt three model stages with a total downsampling ratio of $16$, which differs from the four-stage design with the downsampling ratio of $32$ in the vision transformer model for general vision tasks.  
2) We adopt patch merging layers instead of commonly adopted convolutional layers in transformer and CNN backbones, which introduces less padding on features. 
3) We adopt two types of building layer design, which is different from the coherent transformer layer in most vision transformer variants.  

\begin{figure}[t]
	\centering{\includegraphics[scale = 0.42]{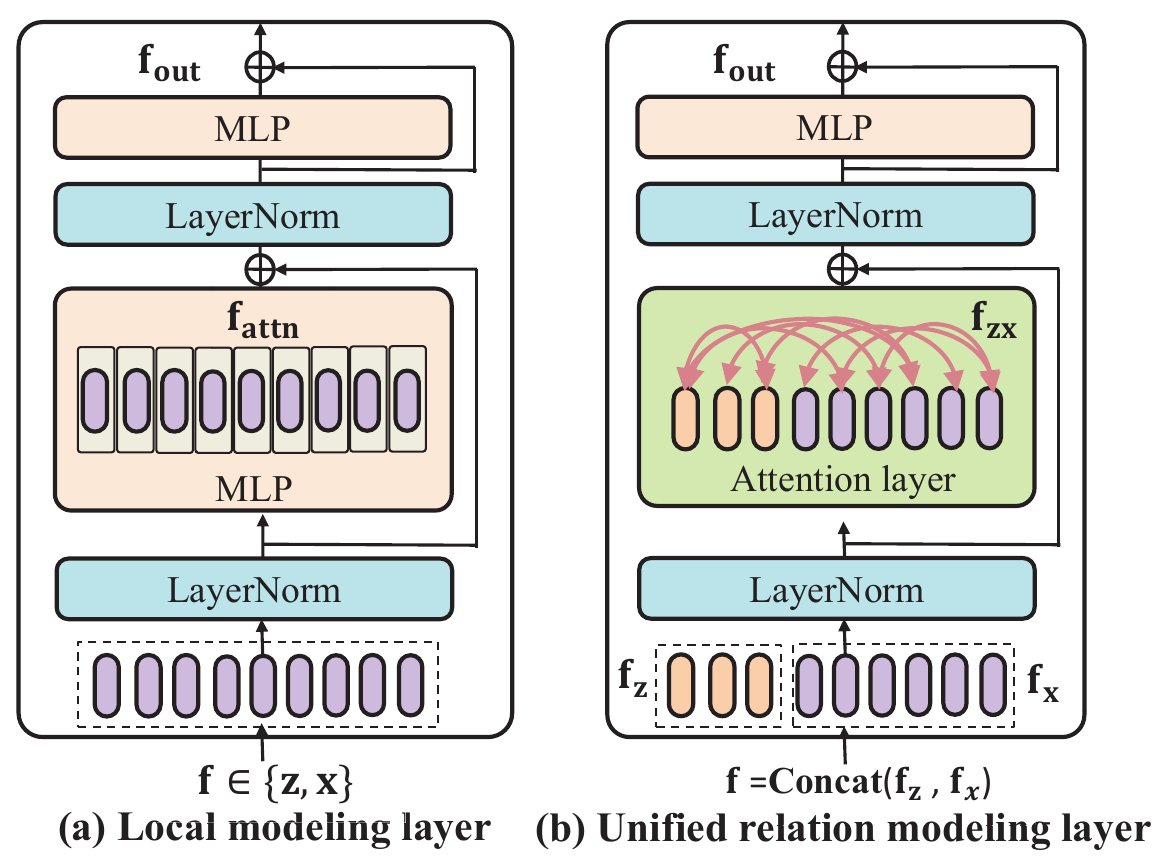}}
	\caption{Detailed architectures of two building layers for SuperSBT. (a) Local modeling layer. (b) Unified relation modeling layer.}  
 \vspace{-4mm}
	\label{fig:grmlm}
\end{figure}

\begin{figure}[t]
	\centering{\includegraphics[scale = 0.28]{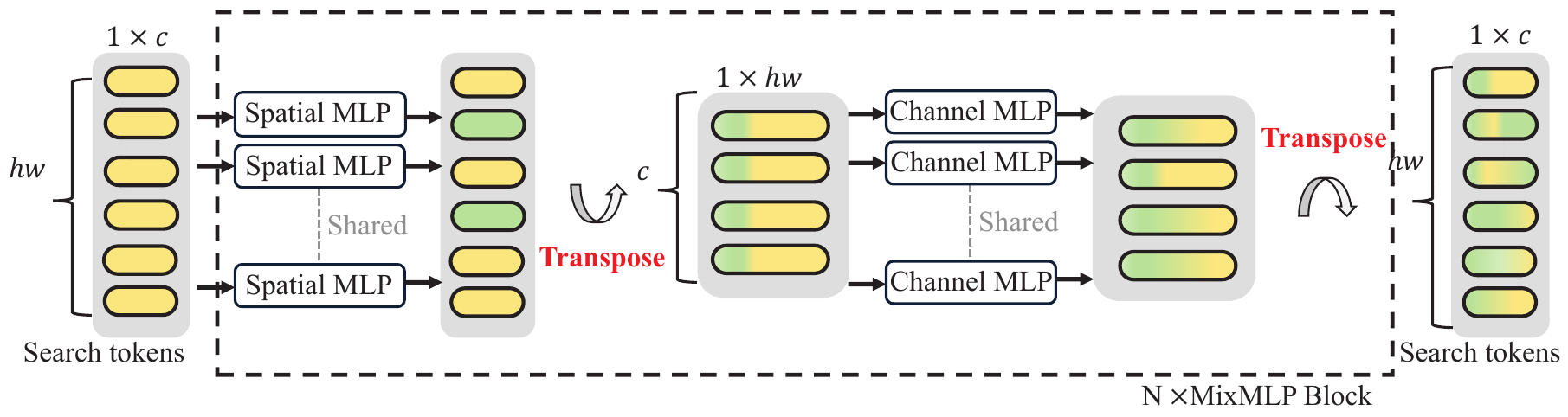}}
	\caption{The detailed architecture of Mix-MLP block for constructing the enhanced prediction head.} 
 \vspace{-2mm}
	\label{fig:mixmlp}
\end{figure}

\begin{figure}[t]
	\centering{\includegraphics[scale = 0.25]{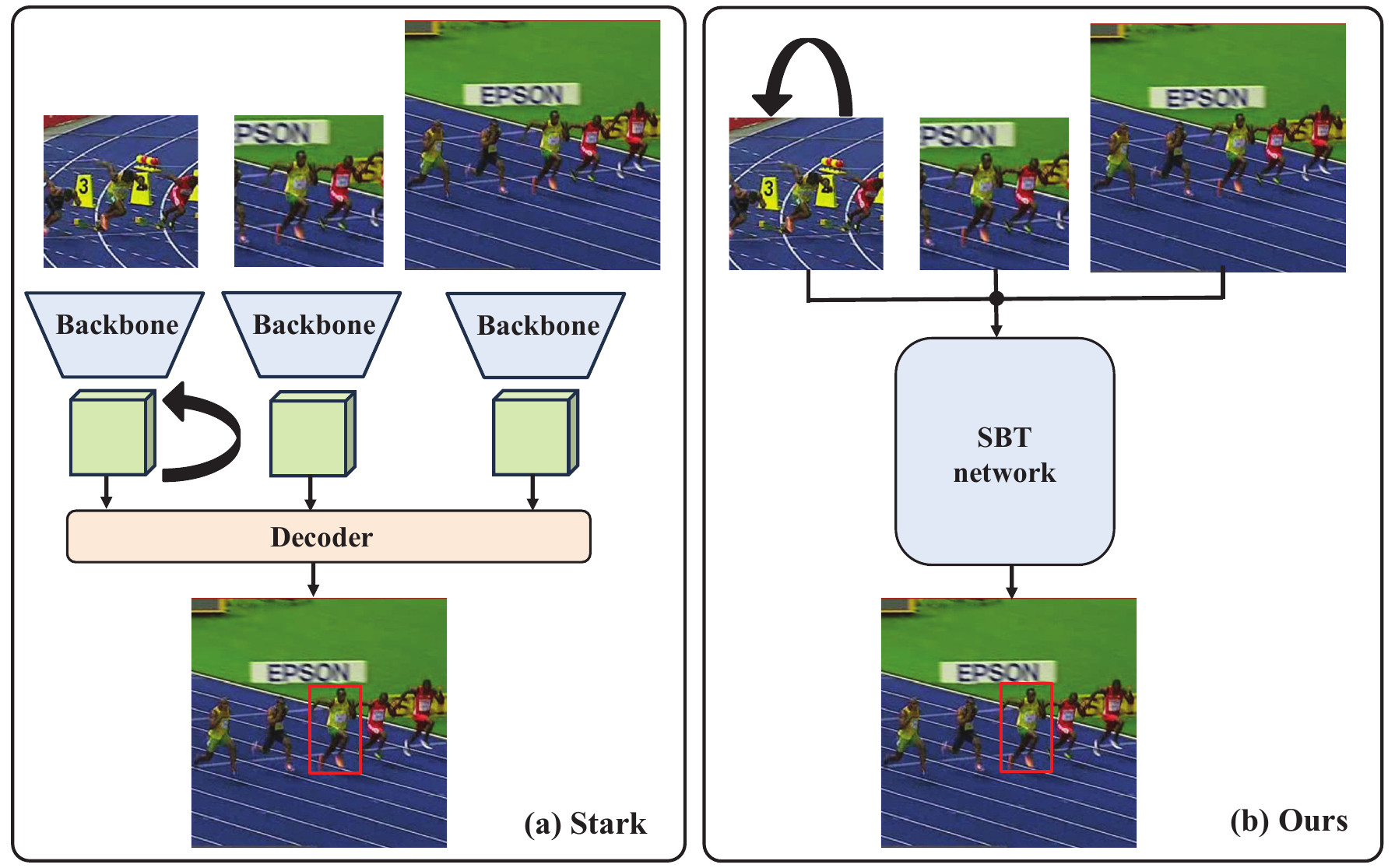}} 
	\caption{{(a) The dynamic template mechanism in Stark~\cite{stark} tracker. (b) Our temporal modeling scheme in SBT tracking.} }
 \vspace{-4mm}
	\label{fig:rv_temporal}
\end{figure}

\subsection{Local Modeling Layer}
\label{sec:lml}
Here, we present the local modeling layer, for the shallow model stage. 
The structure of the local modeling layer is illustrated in Fig.~\ref{fig:grmlm}(a).
Unlike the standard transformer layer\cite{attn, swin, pvt}, the Multi-head attention layer is replaced by the multi-layer perceptron to enrich the local representations in the shallow model stage. 
The remaining part of the local modeling layer follows the standard transformer layer.
The data processing can be formulated as follows:
{\begin{eqnarray}\label{eq:local_mix}
\begin{aligned}
  \mathbf{f}_{att} &= \mathbf{f} + \text{MLP}(\text{LayerNorm}(\mathbf{f}), \quad \mathbf{f}) \in \{z, x\}\\
  \mathbf{f}_{out} &= \mathbf{f}_{att} + \text{MLP}(\text{LayerNorm}(\mathbf{f}_{att})), \quad \mathbf{f} \in \{z, x\}
\end{aligned}
\end{eqnarray}}
where $\mathbf{f}$, $\mathbf{f}_{att}$, and $\mathbf{f}_{out}$ 
are the input, the output of MLP, and the output of the LM layer, respectively.

%
Considering the inability of the attention scheme for local modeling, we adopt a channel-wise convolutional layer in the shallow model stage. 
Observing that shallow features are less expressive, we block the cross-image feature relation modeling. 
Based on the \textbf{Principle E}, we do not place over too many layers whose number is set to $2$ for the shallow model stage.

\subsection{Unified Relation Modeling}
\label{sec:grm}
Here, we illustrate the Unified Relation Modeling (URM) layer for the main stage as shown in Fig.~\ref{fig:grmlm}(b).
Our unified relation modeling layer unifies the intra-image and cross-image relation modeling in a single module. 
The URM layer follows the standard transformer architecture~\cite{attn} and adopts vanilla global attention~\cite{ vit}. 
The template and search image feature tokens are concatenated together without explicitly being divided into cross-image and within-image attention. 
The data processing is as follows:
\begin{eqnarray}\label{eq:global_mix}
\begin{aligned}
  \mathbf{f}_{zx} &= \text{Concat}(\mathbf{f}_{z}, \mathbf{f}_{x}),\\
  \mathbf{f}_{zx-att} &= \mathbf{f}_{zx} + \text{MHSA}(\text{LayerNorm}(\mathbf{f}_{zx})),\\
  \mathbf{f}_{zx-out} &= \mathbf{f}_{zx-att} + \text{MLP}(\text{LayerNorm}(\mathbf{f}_{zx-att})),
\end{aligned}
\end{eqnarray}
where $\mathbf{f}_{zx}$, $\mathbf{f}_{zx-att}$, and $\mathbf{f}_{zx-out}$ are the input, the output of MLP, and the output of the FRM layer, respectively.

%
%
Based on \textbf{Principle B}, we do not adopt any specialized local attention (e.g., window attention), as it is harmful to parallel computing and inference speed.  
Moreover, the URM method can naturally learn the layer pattern of intra-/inter-image modeling for the whole network, which abandons the handcrafted placement of the FRM-SA/-CA layers. 
The information flow in single URM layer $\{ \overrightarrow{zx},\overrightarrow{xz}, \overrightarrow{zz}, \overrightarrow{xx} \}$ is more comprehensive than that in single FRM-SA/-CA layer $\{ \overrightarrow{zz}, \overrightarrow{xx} \}$/$\{ \overrightarrow{zx},\overrightarrow{xz} \}$. 
The computation Flops of attention are calculated as $4C^2L + 2CL^2$, where $L$ is the number of feature tokens. 
The typical hyper-parameters are $\{64, 256\}$ for token numbers $\{ L_{z}, L_{x}\}$, and $C =512$ for channel dimensions in the main stage.
Thus, compared to the FRM layer ($0.34G$), the increased computation overhead of one URM layer ($0.55G$) is around $0.21$G, which has a negligible impact on the tracking efficiency.

\subsection{Relative Position Encoding}
\label{sec:rel}
We use relative position encoding in the main stage's unified relation modeling layers. 
In contrast to the absolute position of each feature token,
this method encodes the relative position between the template and search images into feature tokens.
%
%
Compared to plain ViT~\cite{vit} architecture, which only encodes fixed position information once at the patching embedding layer, 
our relative position encoding can provide better spatial prior for the relation modeling between template and search images in each layer.

\begin{table*}[t]
\newcommand{\splitcell}[1]{\begin{tabular}{@{}c@{}}#1\end{tabular}}
\newcommand{\bsplitcell}[1]{$\left[\splitcell{#1}\right]$}

    \centering
    \renewcommand{\arraystretch}{1.2}
    \renewcommand{\tabcolsep}{2.75 mm}
    \resizebox{0.98\linewidth}{!}{
    \begin{tabular}{c|c|c|c|c|c|c|c} \Xhline{0.4mm}
        Stage & Input Size & Operator & Plain-SBT &Hi-SBT & SuperSBT-Light & SuperSBT-Small & SuperSBT-Base
        \\ \hline
        $0 \rightarrow 1$
        & $128$ \& $256$ 
        & Conv.
        & $K=16$, $S=16$
        & $K=7$, $S=4$
        & \multicolumn{3}{c}{$K=4$, $S=4$}
        \\ \hline
        1 
        & $32$ \& $64$ 
        &\splitcell{MHSA}
        &\bsplitcell{ VG \\$C=768$\\ $H=12$} $\times$ 12
        & \bsplitcell{SRG \\$C=128$\\ $H=1$} $\times$ 3
        & \bsplitcell{MLP \\$C=128$\\ $R=3$} $\times$ 2
        & \bsplitcell{MLP \\$C=128$\\ $R=3$} $\times$ 2
        & \bsplitcell{MLP \\$C=128$\\ $R=3$} $\times$ 2
         \\ \hline
        $1 \rightarrow 2$
        & \textbf{--}
        & \textbf{--}
        & \textbf{--}
        & \textbf{$K=4$, $S=2$}
        &  \multicolumn{3}{c}{Patch Merging ($K=1$, $S=1$)}

        \\ \hline
        2 
        & $16$ \& $32$
        & \splitcell{MHSA}
        & \textbf{--}
        & \bsplitcell{SRG \\$C=256$\\ $H=12$} $\times$ 4
        & \bsplitcell{MLP \\$C=256$\\ $R=3$} $\times$ 2
        & \bsplitcell{MLP \\$C=256$\\ $R=3$} $\times$ 2
        & \bsplitcell{MLP \\$C=256$\\ $R=3$} $\times$ 2
           \\ \hline
                $2 \rightarrow 3$
        & \textbf{--}
        & \textbf{--}
        & \textbf{--}
        & \textbf{$K=4$, $S=2$}
        &  \multicolumn{3}{c}{Patch Merging  ($K=1$, $S=1$)}
        \\ \hline
        3 
        & $8 \& 16$
        & \splitcell{MHSA }
        &  \textbf{--}
        & \bsplitcell{SRG \\$C=320$\\ $H=12$} $\times$ 10
        & \bsplitcell{VG \\$C=512$\\ $H=8$} $\times$ 6
        & \bsplitcell{VG \\$C=512$\\ $H=8$} $\times$ 10
        & \bsplitcell{VG \\$C=512$\\ $H=8$} $\times$ 20
        
        \\ \hline
        Head
        & $1\times 1$
        & \textbf{--}
        & \multicolumn{2}{c|}{Conv-BN-Relu $\times$3 }
                & \multicolumn{3}{c}{Mix-MLP block $\times$3 }

        \\ \hline
        \multicolumn{3}{c|}{\#Params} & 86.7M & 21.2M & 21.5 M & 34.3M & 65.5M
        \\ \hline
        \multicolumn{3}{c|}{\#Flops}  & 27.4 G & 10.2G & 10.4G & 14.5G & 24.6 G
        \\ \hline
        \multicolumn{3}{c|}{\#Speed} 
        & 92 FPS 
        & 37 FPS 
        & 141 FPS  
        & 110 FPS  
        & 81 FPS 
        \\ 
        \Xhline{0.4mm}
    \end{tabular}
    }
    \caption{\textbf{Detailed settings of the proposed SBT and SuperSBT.}
    The parameters of building layers are shown in brackets,  with the number of layers stacked.
    Plain-SBT adopts plain ViT architecture, while Hi-SBT adopts a hierarchical structure and spatial-reduction attention. 
    Three improved SBT models (SuperSBT) of different model scales are presented, including light, small, and base versions
    %
    }
\label{tab:sbt_variant}
 \vspace{-4mm}

\end{table*}

\subsection{Masked Image Modeling Pre-training}
\label{sec:mim}
%
%
We follow the masked image modeling pre-training method in~\cite{mae} to pre-train our model in ImageNet~\cite{ImageNet}.
A large random subset of image patches (e.g., 75\%) is masked during pre-training. 
The encoder, adopted as the SuperSBT network, is applied to the small subset of visible patches. 
Mask tokens are added after the encoder, and the full set of encoded patches and mask tokens is processed by a small decoder that reconstructs the original image in pixels. 
After pre-training, the decoder is discarded, and the encoder is applied to fine-tune the tracking task.

%
Our SuperSBT network is modified to better adapt to masked image modeling pre-training:
1) SuperSBT adopts a vanilla global attention operator because non-global operations (e.g., window attentions) hinder the hierarchical models from determining whether each pair of tokens needs to communicate (e.g., window attentions).
2) Stage 4 is removed so that all the tokens in Stage 3 are symmetric; we can directly discard all masked patches from the input. Thus, our SuperSBT can be adopted directly as the encoder. 

\subsection{Temporal Modeling Scheme}
\label{sec:temporal}
{We demonstrate our temporal modeling scheme to capture the target appearance changes over time. 
Similar to the dynamic template mechanism in Stark~\cite{stark},
we introduce a dynamically updated template sampled from intermediate frames as an additional input, as shown in Fig.~\ref{fig:rv_temporal}(b). 
In the main stage of SueprSBT, triplet image inputs (first frame, dynamic template, and search frame) are concatenated and then sent to the FRM layers.
The dynamic template is initialized from the first frame and updated by a predefined confidence threshold and time interval.
More details are in the supplementary materials.}

Our SuperSBT can directly update the image input while Stark updates the image features extracted from ResNet~\cite{resnet} backbone, and only the decoder part models the temporal variations. Thus, our SuperSBT learns the spatiotemporal relationship among templates and search regions more thoroughly than Stark.

%
%

\subsection{Mix-MLP Prediction Head}
\label{sec:mixmlp}
In addition to the Convoluational-based head, we propose a Mix-MLP head, which can jointly model the dependency between the spatial and channel dimensions of the input features.
{The detailed analysis is in supplementary materials.}

\paragraph{Mix-MLP head.}
Fig.~\ref{fig:mixmlp} shows the architecture of Mix-MLP blocks. 
We implement regression head $\Phi_{reg}$ and classification head $\Phi_{cls}$ by stacking multiple Mix-MLP blocks. 
Inside each block, we drop the template tokens and feed the search image tokens:
\begin{equation}
\begin{aligned}
&\hat{f}^{i}=  \varphi_{sp} ({\rm RS}( \varphi_{cn} ({\rm RS}(\hat{f}^{i-1})))), \\
\end{aligned}
\end{equation}
where $\varphi_{sp}$ and $\varphi_{cn}$ consist of a linear layer followed by RELU activation. $\rm RS$ represents reshape. $\varphi_{cn}$ is applied to features along the channel dimension, and the weights are shared for all spatial locations. In contrast, the operator $\varphi_{sp}$ is shared for all channels.

\subsection{Network Variants}
\label{sec:variant}
Following the guidelines from Sec.~\ref{sec:empirical}, five variants of SBT are described in Tab.~\ref{tab:sbt_variant}.
It includes two base SBT variants and three SuperSBT variants.
We place most layers in the third model stage based on \textbf{Principle C}.
Moreover, we adjust the channel numbers and layer dimensions to maintain high speed and model performance based on \textbf{Principle E}.

\subsection{Training Loss}
\label{subsec:loss}
During training, we reshape the sequence of search region tokens to a 2D spatial feature map and then feed it into the tracking head.
The tracking head is decomposed into three subtasks:
a classification branch to predict the coarse location of the target center, a local offset branch to compensate for the target center, and a target size branch to predict the target's normalized bounding box (i.e., width and height). 
Thus, we have three kinds of outputs from the tracking head: the target position score map $ A^{score}_{w \times h \times 1} \in [0,1]$, local offset map $ A^{local}_{w \times h \times 2} \in [0,1] $ and target size regression map $ A^{reg}_{w \times h \times 2} \in [0,1] $.  
Instead of optimizing three subtasks independently~\cite{SiamCAR, siamrpn++}, 
we first localize the target coarsely from the target position score map, then obtain the accurate location and target size by local offset map and target size regression map.
To be specific, the position with the highest classification score is considered to be the coarse target position:
\begin{equation}
(x_c, y_c) =\arg\max_{i,j}\{A^{score}_{w \times h \times 1}(i,j,:)\},
\label{getq}
\end{equation}
where $(x_c, y_c)$ is the normalized coordinate of the target center. 
Then, we obtain the compensation of the target center and size estimation at the corresponding location $(x_c, y_c)$.
Thus, the finial target bounding box $(x, y, w, h)$ is calculated as:
\begin{equation}
\begin{split}
&(x, y)=(x_c, y_c)+ A^{local}_{w \times h \times 2}(x_c,y_c,:), \\
&(w, h)= A^{reg}_{w \times h \times 2}(x_c,y_c,:). \\
\end{split}
\end{equation}
During training, we adopt the weighted focal loss~\cite{Gfocalloss} for the classification branch. 
Specifically, for each ground truth target center  $\hat{p}$ and its corresponding low-resolution equivalent  $\tilde{p}=[\tilde{p}_x, \tilde{p}_y]$, the ground truth heatmap can be generated using a Gaussian kernel as $\hat{\text{P}}_{xy}=\exp \left(-\frac{\left(x-\tilde{p}_{x}\right)^{2}+\left(y-\tilde{p}_{y}\right)^{2}}{2 \sigma_{p}^{2}}\right)$, where $\sigma$ is an object size-adaptive standard deviation~\cite{cornernet}. 
%

Finally, with the predicted bounding box, L1 loss, and the generalized IoU loss are employed for bounding box regression. 
The overall loss function is as follows:
\begin{equation}
	\label{equ-loss-loc}
	\begin{aligned}
		L_{overall}=L_{cls} + \lambda_{iou}L_{iou} + \lambda_{L_1}L_1,
	\end{aligned}
\end{equation}
where $\lambda_{iou}=2$ and $\lambda_{L_1}=5$ are the regularization parameters in our experiments as in~\cite{stark}.

\begin{table*}[!t]
\begin{center}
\resizebox{0.96\linewidth}{!}{
  \setlength{\tabcolsep}{3mm}{  
  \small
\begin{tabular}{l|c|ccc c ccc c ccc c cc c}
\toprule
\multirow{2}{*}{Method} & \multirow{2}{*}{Year} &\multicolumn{3}{c}{LaSOT~\cite{LaSOT}}	
&
&\multicolumn{3}{c}{TrackingNet~\cite{trackingnet}}
&
&\multicolumn{3}{c}{GOT-10k~\cite{got}}
&
&\multicolumn{3}{c}{LaSOT-Ext~\cite{lasot_ext}}\\

    \cline{3-5}
    \cline{7-9}
    \cline{11-13}
        \cline{15-16}

& &AUC	&P$_{Norm}$	&P	& &AUC	&P$_{Norm}$	&P &	&AO	&SR$_{0.5}$	&SR$_{0.75}$ &     &AUC	&P\\
\midrule[0.5pt]

SuperSBT-Light 
&ours	
& 65.8
& 75.3
& 70.6&

& 81.4
& 86.2
& 79.3

&&	69.4
& 79.4
& 64.1 
&
& 44.9
& 50.1\\
SuperSBT-Small  &ours	
&67.5
&77.1
&73.1&

& 82.7
& 87.3
& 81.2

&&71.6	
&81.4
&68.3
&
& 45.2
& 50.3\\

SuperSBT-Base  &ours
& \textcolor{blue}{\textbf{70.0}}
& \textcolor{blue}{\textbf{79.8}}
& \textcolor{cGreen}{\textbf{76.1}}
&

& \textcolor{blue}{\textbf{84.0}}
&\textcolor{cGreen}{\textbf{88.4}}	
&\textcolor{cGreen}{\textbf{83.2}}	
&
&\textcolor{cGreen}{\textbf{74.4}}	
&\textcolor{cGreen}{\textbf{83.9}}
&\textcolor{cGreen}{\textbf{71.3}}
&
& 48.1
& 54.2\\

SuperSBT-Base-384  &ours
& \textcolor{red}{\textbf{72.8}}
& \textcolor{red}{\textbf{82.5}}
& \textcolor{red}{\textbf{78.6}}
&

& \textcolor{red}{\textbf{84.8}}
&\textcolor{red}{\textbf{88.9}}	
&\textcolor{red}{\textbf{83.7}}	
&
&\textcolor{red}{\textbf{75.5}}	
&\textcolor{red}{\textbf{84.3}}
&\textcolor{red}{\textbf{72.4}}
&
& \textcolor{red}{\textbf{50.7}}
& \textcolor{red}{\textbf{57.9}}\\

\midrule[0.1pt]
Plain-SBT &ours 
&65.7	&75.4	& 70.1 
&	
&81.2	& 86.1 &80.3	& 
&69.2	& 79.1	& 64.0
&
& 44.2
& 49.3
\\
Hi-SBT &ours 
&65.3	&74.7	&70.3
&	
&81.0	&85.6	&79.0
& 
&69.5 &79.8	&63.9
& 
& 44.6
& 49.6

\\
\midrule[0.1pt]
SeqTrack~\cite{seqtrack}	&2023	
& \textcolor{cGreen}{\textbf{69.9}}	
& \textcolor{cGreen}{\textbf{79.7}}
& \textcolor{blue}{\textbf{76.3}}
&
& 83.3
& 88.3
& 82.2
&
& \textcolor{blue}{\textbf{74.7}}
& \textcolor{blue}{\textbf{84.7}}
& \textcolor{blue}{\textbf{71.8}}
&
&  \textcolor{blue}{\textbf{49.5}}
&  \textcolor{red}{\textbf{56.3}}
\\
ROMTrack~\cite{romtrack}	&2023	
& 69.3	
& 78.8
& 75.6
&
& 83.6
& 88.4
& 82.7
&
& 72.9
& 82.9
& 70.2
&
& 48.9
& 55.0
\\
GRM~\cite{grm}	&2023	
& 69.9
& 79.3
& 75.8
&
&\textcolor{cGreen}{\textbf{84.0}}
&\textcolor{blue}{\textbf{88.7}}
&\textcolor{blue}{\textbf{83.3}}
&
&73.4
&82.9
&70.4
&
& -
& -
\\
OStrack~\cite{ostrack}	&2022	
&	69.1
& 78.7
& 75.2
&
& 83.1
& 87.8
& 82.0
&
& 71.0
& 80.4
& 68.2
&
& 47.4
& 53.3
\\
SimTrack~\cite{simtrack}	&2022	
&69.3
& 78.5
& -
&
&82.3
& 86.5
&-	
&
&68.6
&78.9 
&62.4
&
&-	
&-	

\\
Mixformer~\cite{mixformer}	&2022	
&69.2	
& 78.7
& 74.7
&
&83.1
&88.1
&81.6
&
&70.7
&80.0
&67.8
&
& -
& -
\\
SwinTrack~\cite{swintrack}	&2022	
&67.2	
&-
& 70.8
&
&81.1
&-
&78.4	
&
&71.3
&81.9
&64.5
&
&  \textcolor{cGreen}{\textbf{49.1}}
&  \textcolor{cGreen}{\textbf{55.6}}
\\

TransT~\cite{transt}	&2022	
&64.9	
&73.8
&69.0
&
&81.4
&86.7
&80.3
&
&67.1
&76.8
&60.9
&
& -
& -	

\\

ToMP50~\cite{tomp}	&2022 &67.6 &-	&- &	&81.2 	&86.4	&78.9 &	&-	&-	&-&
& 45.9
& -\\
CSWinTT~\cite{cswint} &2022	&66.2 &75.2	&70.9 &	&81.9	&86.7	&79.5	& &69.4	&78.9	&65.4&
& -
& -\\
UTT~\cite{UTT} &2022	&64.6 &-	&67.2 &	&79.7	&-	&77.0 &	&67.2	&76.3	&60.5&
& -
& -\\
ARDiMPsuper~\cite{AlphaRefine} &2021	&65.3	&73.2	&68.0 &	&80.5	&85.6	&78.3 &	&70.1	&80.0	&64.2&
& -
& -\\
TrDiMP~\cite{tmt} &2021	&63.9	&-	&61.4 &	&78.4	&83.3	&73.1	&	&68.8	&80.5	&59.7&
& -
& -\\
TrSiam~\cite{tmt}  &2021	&62.4	&-	&60.0	& &78.1	&82.9	&72.7	&	&67.3	&78.7	&58.6&
& -
& -\\
STMTrack~\cite{stmtrack}  &2021	&60.6	&69.3	&63.3 &	&80.3 &85.1	&76.7	&		&64.2	&73.7	&57.5&
& -
& -\\
SiamBAN-ACM~\cite{siamban} &2021 	&57.2	&-	&- &	&75.3	&81.0	&71.2	&	&-	&-	&-&
& -
& -\\
SiamGAT~\cite{siamgat}  &2021	&53.9	&63.3	&53.0 &	&-	&-	&-	&	&62.7	&74.3	&48.8&
& -
& -\\
DSTrpn~\cite{DSTrpn}  &2021	&43.4	&51.3	&- &	&64.9	&58.9	&-	&	&-	&-	&-&
& -
& -\\
SiamR-CNN~\cite{siamrcnn}   &2020	&64.8	&72.2	&- &	&81.2	&85.4	&80.0	&	&64.9	&72.8	&59.7&
& -
& -\\
Ocean~\cite{Ocean}	&2020	&56.0	&65.1 	&56.6 &	&-	&- 	&- &	&61.1	&72.1	&47.3&
& -
& -\\
KYS~\cite{kys}	    &2020	&55.4	&63.3	&- &	&74.0	&80.0	&68.8 &	&63.6	&75.1	&51.5&
& -
& -\\
DCFST~\cite{dcfst}	&2020	&-	&-	&- &	&75.2	&80.9	&70.0 &	&63.8	&75.3	&49.8&
& -
& -\\
SiamFC++~\cite{siamfcpp}	&2020	&54.4	&62.3	&54.7 &	&75.4	&80.0	&70.5 &	&59.5	&69.5	&47.9&
& -
& -\\
PrDiMP~\cite{PrDiMP} &2020	&59.8	&68.8	&60.8 &	&75.8	&81.6	&70.4 &	&63.4	&73.8	&54.3&
& -
& -\\
SiamAttn~\cite{attnsiam}	&2020	&56.0	&64.8	&- &	&75.2	&81.7	&- &	&-	&-	&-&
& -
& -\\
MAML~\cite{MAML}	&2020	&52.3	&-	&- &	&75.7 	&82.2	&72.5 &	&-	&-	&-&
& -
& -\\
D3S~\cite{D3S}	    &2020	  &-	&-	&- &	&72.8	&76.8	&66.4 &	&59.7	&67.6	&46.2&
& -
& -\\
SiamCAR~\cite{SiamCAR}	&2020	  &50.7	&60.0	&51.0 &	&-	&-	&-	 & &56.9	&67.0 	&41.5&
& -
& -\\
SiamBAN~\cite{siamban}	&2020	&51.4	&59.8	&52.1 &	&-	&-	&- &	&-	&-	&-&
& -
& -\\
DiMP~\cite{DiMP}	&2019    	&56.9	&65.0	&56.7 &	&74.0	&80.1	&68.7 &	&61.1	&71.7	&49.2&
& 39.2
& 45.1\\
SiamPRN++~\cite{siamrpn++} &2019	&49.6	&56.9	&49.1 &	&73.3	&80.0	&69.4 &	&51.7	&61.6	&32.5&
& 34.0
& 39.6\\
ATOM~\cite{atom}	  &2019  	&51.5	&57.6	&50.5 &	&70.3	&77.1	&64.8 &	&55.6	&63.4	&40.2&
& 37.6
& 43.0\\
ECO~\cite{ECO}	      &2017  	&32.4	&33.8	&30.1 &	&55.4	&61.8	&49.2 &	&31.6	
&11.1\\
MDNet~\cite{MDnet}	 &2016   	 &39.7	&46.0	&37.3 &	&60.6	&70.5	&56.5 &	&29.9	&30.3	&9.9&
& 27.9
& 31.8\\
SiamFC~\cite{siamfc}	&2016	&33.6	&42.0	&33.9 &	&57.1	&66.3	&53.3 &	&34.8	&35.3	&9.8&
& -
& -\\
 \bottomrule
\end{tabular}}}
\caption{{State-of-the-art comparison on LaSOT, TrackingNet, and GOT-10k. The best three are shown in \textbf{\textcolor{red}{red}}, \textbf{\textcolor{blue}{blue}}, and \textbf{\textcolor{cGreen}{green}} fonts.}}
\vspace{-4mm}
\label{tab:sota1}
\end{center}
\end{table*}

\section{Experiments}
\label{sec:experiment}

This section describes the implementation details of our method, performance comparisons with the state-of-the-art (SOTA) trackers, and ablation studies.

\subsection{Implementation Details}

{\noindent \textbf{Offline training. }}
To train SBT and SuperSBT, our model undergoes training on the training splits of COCO~\cite{COCO}, TrackingNet~\cite{trackingnet}, LaSOT~\cite{LaSOT}, and GOT-10k~\cite{got} datasets. We extract image pairs from individual video sequences to create training samples for video datasets (TrackingNet, LaSOT, and GOT-10k). In the case of COCO detection datasets, we introduce certain transformations to the original images to produce image pairs. Standard data augmentation techniques, including translation and brightness jitter, are applied to augment the training set.
Unless otherwise specified, the search region patch and template patch scales are set to  $256 \times 256$ and $128 \times 128$, respectively. 
For SuperSBT-384, the corresponding scales are set to $384$ and $192$.
The backbone parameters are initialized using a masked image modeling pre-trained weights from ImageNet\cite{ImageNet}, while other parameters of our model are initialized with Xavier~\cite{xaiv} initialization. Training is conducted using AdamW~\cite{AdamW}, with the learning rate for the backbone set to 1e-5, that for other parameters to 1e-4, and weight decay to 1e-4.
The training is conducted on eight Tesla V100 GPUs, each handling 16 sample pairs (i.e., a total batch size of 128). The training spans 500 epochs, with 60,000 sample pairs processed in each epoch. The learning rate is reduced by a factor of 10 after 400 epochs.


To train the dynamic template modeling of SuperSBT,  we select the head frame or the tail frame to generate the search region. 
We generate the initial template patch using the frame farthest from the search region frame, and the updated template patches are generated using the middle frames. In the case of the COCO detection dataset, we apply transformations to create image pairs. Specifically, we randomly choose one image to form the search region patch, another for the initial template patch, and the remaining images for the updated template patches. To simulate inaccuracies in template updating during tracking, we introduce jitter and transformations to the position and shape of the target box for the search region patch and the updated templates. No transformations or jittering are applied initially, as the provided bounding box is typically accurate in online tracking.

{\noindent \textbf{Online tracking. }}
The prediction head generates multiple box candidates and their confidence scores during online tracking. Subsequently, postprocessing of these scores involves applying a window penalty. More precisely, a Hanning window is employed on the scores.
The window penalty penalizes the confidence scores of feature points distant from the target in previous frames.
Ultimately, we choose the box with the highest confidence score as the tracking outcome.
In the case of SBT and SuperSBT, we consistently employ the initial bounding box of the first frame as the template and refrain from updating it.
On the other hand, for the dynamic template version of SuperSBT, two templates are utilized by default, and the initial template is retained while other templates are subject to updates. The maximum score governs the update of the dynamic template -- if the score surpasses the update threshold, the dynamic template will be updated using the predicted result.

\subsection{Evaluation on TrackingNet, LaSOT, and GOT-10k}

In this subsection, we compare our methods with state-of-the-art trackers on the large-scale LaSOT~\cite{LaSOT}, TrackingNet~\cite{trackingnet}, and GOT-10k~\cite{got} datasets. The results are shown in Tab.~\ref{tab:sota1}.

{\noindent \textbf{LaSOT. }} LaSOT~\cite{LaSOT} is a recent large-scale dataset that contains 1,400 challenging videos: 1,120 for training and 280 for testing.
We follow the one-pass evaluation (Success and Precision) to compare different tracking algorithms on the
LaSOT test set.
As shown in Tab.~\ref{tab:sota1},
SuperSBT-Base-384 achieves the best results.
Specifically, our SuperSBT-Base method outperforms two fully transformer-based tackers, ROMTrack~\cite{romtrack} and OStrack~\cite{ostrack}, by 0.7\% and 0.9\% higher, respectively, in terms of AUC.
Compared with hybrid CNN+transformer trackers, our SuperSBT-Base outperforms TransT~\cite{transt} and SwinTrack~\cite{swintrack} by large margins of 5.1\% and 2.3\%, respectively.
%

\begin{table}[t]\normalsize
  \centering
\resizebox{0.95\linewidth}{!}{
  \setlength{\tabcolsep}{2mm}{
    \small
    \begin{tabular}{l|cccc}
    \toprule
    Method &NFS~\cite{NFS}&OTB~\cite{otb}&UAV123~\cite{UAV}&TNL2k~\cite{tnl2k}\\
    \midrule[0.5pt]
SuperSBT-Base & \textbf{\textcolor{red}{67.1}}&68.9& \textbf{\textcolor{red}{69.5}}& \textbf{\textcolor{red}{56.6}}\\
SuperSBT-Small &66.2 &68.0& \textbf{\textcolor{cGreen}{68.8}}& \textbf{\textcolor{blue}{ 55.7}}\\
SuperSBT-Light &65.1 &66.1& 67.3&\textbf{\textcolor{cGreen}{53.6}}\\

    \midrule[0.1pt]
    TransT &\textbf{\textcolor{cGreen}{65.7}}&\textbf{\textcolor{cGreen}{69.4}}&\textbf{\textcolor{blue}{69.1}}
    &50.7\\
    TrDiMP~\cite{tmt} &\textbf{\textcolor{blue}{66.5}}&\textbf{\textcolor{blue}{71.9}}&67.5&-\\
    SiamBAN-ACM\cite{siamban} &-&\textbf{\textcolor{red}{72.0}}&64.8&-\\
    DiMP\cite{DiMP} &62.0&68.4&65.3&44.7\\
    SiamRPN++\cite{siamrpn++} &50.2&69.6&61.3&39.8\\
    ATOM\cite{atom} &58.4&66.9&64.2&40.1\\
    ECO\cite{ECO} &46.6&69.1&53.2&32.6\\
    MDNet\cite{MDnet} &42.2&67.8&52.8&-\\
    \bottomrule
    \end{tabular}
    }
    
  }
\caption{{Comparison with the state-of-the-arts on the NFS, OTB, TNL2k, and UAV123 datasets in terms of AUC score. The best three results 
are shown in \textbf{\textcolor{red}{red}}, \textbf{\textcolor{blue}{blue}}, and \textbf{\textcolor{cGreen}{green}} fonts.}}
\vspace{-3mm}
\label{tab-sota-small}
\end{table}

{\noindent \textbf{TrackingNet.}} TrackingNet~\cite{trackingnet} is a large-scale tracking dataset that covers diverse object classes and scenes.
Its test set contains 511 sequences of publicly available ground truth.
We submit our tracker's outputs to the official online evaluation server and report the Success (AUC) and Precision (P and P$_{Norm}$) results in Tab.~\ref{tab:sota1}.
As can be seen, SuperSBT-Base-384 achieves the best performance, achieving 84.0\%, 88.4\%, and 83.2\% in terms of AUC, $\rm{P}_{Norms}$, and P, respectively. Besides, in terms of AUC, SuperSBT-Base surpasses the fully-transformer tracker OStrack~\cite{ostrack} by 0.9\% and two transformer-based trackers, TrDiMP and TrSiam, by  5.6\% and 5.9\%, respectively.

{\noindent \textbf{GOT-10k.}} The GOT-10k~\cite{got} dataset contains 10k sequences for training and 180 for testing.
We follow the defined protocol presented in~\cite{got} and submit the tracking outputs to the official evaluation server.
The results (i.e., AO and SR$_{\rm{T}}$) are reported in Tab.~\ref{tab:sota1}.
It can be seen that our SuperSBT-Base-384 achieves the best performance. SuperSBT-Base and SuperSBT-Light obtain 74.4\% and 69.4\% AO, respectively, leading previous methods by significant margins. 
SuperSBT-Base achieves comparable performance in terms of AO at a much faster speed compared to SeqTrack~\cite{seqtrack} under the same input resolution (81 FPS vs 37 FPS).
Furthermore, SuperSBT-Base outperforms the recent transformer tracker GRM~\cite{grm} by 1.0\% in terms of AO while running at a faster speed (81 FPS vs 40 FPS).

\noindent \textbf{{LaSOT-Ext.}} LaSOT-Ext~\cite{lasot_ext} is a recently released dataset with 150 video sequences and 15 object classes, which contains highly challenging scenarios. Our SuperSBT-Base still outperforms the SOTA trackers OStrack and achieves the top rank with $50.7\%$ AO with the $384$ input setting.
Our small variant achieves comparable performance to ToMP50 ($45.2 \%$ vs. $45.9 \% $) while running much faster (110 FPS vs. 30 FPS).
It validates the effectiveness and efficiency of our method.

\begin{figure}[!t]
	\centering{\includegraphics[scale = 0.17]{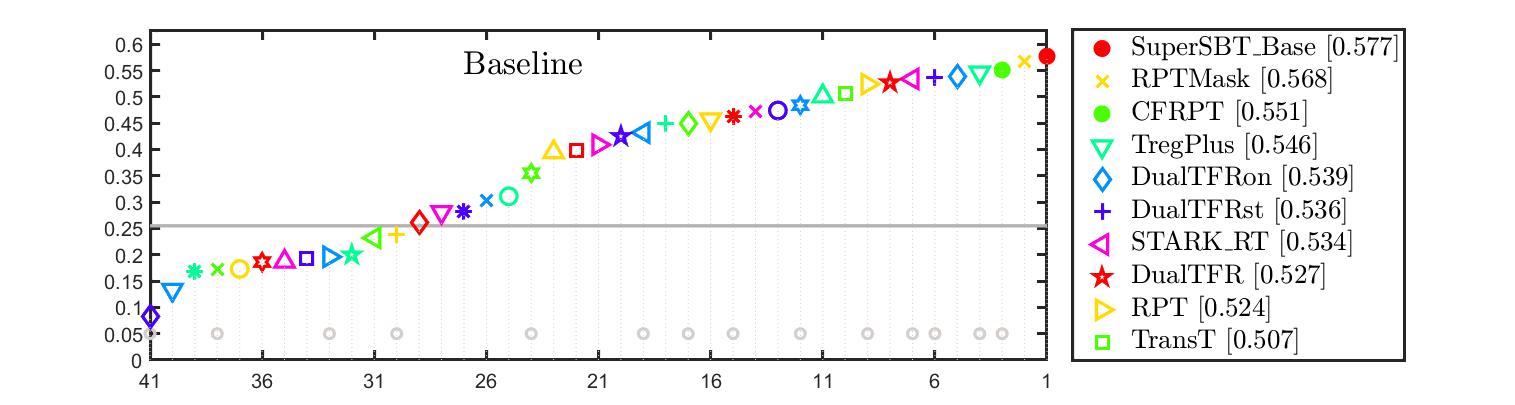}}
	\caption{EAO rank plots on VOT2021 }
	\label{fig:eao2021}

\end{figure}

\begin{table}[t]
	\centering
    \setlength\tabcolsep{2.5pt}

    \resizebox{\linewidth}{!}{
    	\begin{tabular}{ccccccc|c}
            \Xhline{1.0pt}
    		\multirow{2}*{Case} &\multirow{2}*{Siamese} & \multirow{2}*{Backbone} & \multirow{2}*{Operator} & \multicolumn{3}{c|}{Feature utilization} & \multicolumn{1}{c}{GOT-10k} \\
    	~ &	~ & ~ & ~ & Low & Mid  & High & AO   \\
      
            \Xhline{1.0pt}
            \ding{172} & \CheckmarkBold & CNN  & DWC &S2 &S3 &S4  & \textbf{56.2}    \\
            \ding{173} &\CheckmarkBold & CNN & CA &S2 &S3 &S4 & \textbf{57.5}    \\
            
    	\ding{174} & \CheckmarkBold & CNN & DCF  &- &- &S4  & \textbf{30.3}   \\
           \ding{175} & \CheckmarkBold & Trans. & DWC &L6 &L8 &L10 & \textbf{60.1}    \\
           \ding{176} & \CheckmarkBold & Trans.& CA&L6 &L8 &L10  & \textbf{61.5}  \\
                		\hline
           \ding{177} & \XSolidBrush&Trans.  &DCF &-&- &L10  & \textbf{31.5}   \\
            \rowcolor{gray!20}  \ding{178} & \XSolidBrush&Trans.  &- &L6 &L8 &L10  & \textbf{65.0}   \\
          \ding{179} &  \XSolidBrush&Trans.  &DWC &L6 &L8 &L10  &\textbf{65.9}    \\
           \ding{180} & \XSolidBrush&Trans.  &DCF &L6 &L8 &L10  & \textbf{35.2}   \\
            \Xhline{1.0pt}
    	\end{tabular}
    }
        \caption{\textbf{Alignment comparison with the Siamese tracking pipeline.} Ablation studies are conducted on GOT-10k~\cite{got}. 
    S3L6 denotes $6^{th}$ layer within the $3^{rd}$ stage.
    Trans. denotes our full transformer-based backbone. 
    The correlation operator includes DWC (depth-wise correlation~\cite{siamrpn++}), CA, and DCF~\cite{atom}.
    }
    \label{tab:ablation_structure}
\end{table}

\begin{table}[t]
\begin{center}
\resizebox{1.0\linewidth}{!}{
  \setlength{\tabcolsep}{3mm}{  
  \small
\begin{tabular}{l|ccc c ccc c c c}
\toprule
\multirow{2}{*}{Method} &\multicolumn{3}{c}{LaSOT~\cite{LaSOT}}	

&\multicolumn{3}{c}{GOT-10k~\cite{got}}
&
&\multicolumn{1}{c}{NFS~\cite{NFS}}
& Speed in GPU
\\
    \cline{2-4}
    \cline{6-8}
&AUC	&P$_{Norm}$	&P	& &AO	&SR$_{0.5}$	&SR$_{0.75}$  &AUC	 &FPS	 \\
\midrule[0.5pt]

SuperSBT-Light 
& \textcolor{blue}{\textbf{65.8}}
& \textcolor{blue}{\textbf{75.3}}
& \textcolor{blue}{\textbf{70.6}}
&

& \textcolor{blue}{\textbf{69.4}}
& \textcolor{blue}{\textbf{79.4}}
& \textcolor{blue}{\textbf{64.1 }}

& \textcolor{blue}{\textbf{65.1}}
& 141

\\
SuperSBT-Small 	
&\textcolor{red}{\textbf{67.5}}
&\textcolor{red}{\textbf{77.1}}
&\textcolor{red}{\textbf{73.1}}
&

&\textcolor{red}{\textbf{71.6}}
&\textcolor{red}{\textbf{81.4}}
&\textcolor{red}{\textbf{68.3}}

& \textcolor{red}{\textbf{66.2}}
& 110

\\
\midrule[0.1pt]

HiT-Tiny~\cite{hit}	 &54.8&60.5&52.9 &	&52.6 	&59.3	&42.7 	&53.2 &204
\\
HiT-Small~\cite{hit}	 &60.5 &68.3	&61.5 &	&62.6 	&71.2	&54.4 	&61.8	 &192
\\
HiT-Base~\cite{hit} &\textcolor{cGreen}{\textbf{64.6}} &\textcolor{cGreen}{\textbf{73.3}}	&\textcolor{cGreen}{\textbf{68.1}} &	&\textcolor{cGreen}{\textbf{64.0}} &\textcolor{cGreen}{\textbf{72.1}}	&\textcolor{cGreen}{\textbf{58.1}} 	&\textcolor{cGreen}{\textbf{63.6}} &175
\\
E.T.Track~\cite{ettrack} 	&59.1 &-	&- &	&- 	&-	&- 	&59.0	 &-\\
FEAR-XS~\cite{fear}	&53.5 &-	&54.5 &	&61.9 	&-	&- 	&61.4	 &-\\
FEAR-M~\cite{fear} 	&54.6 &-	&55.6 &	&62.3 	&-	&- 	&62.2	 &-\\

LightTrack-Mobile~\cite{lighttrack}	&53.8 &-	&53.7 &	&61.1 	&71.0	&- 		&- & -
\\

LightTrack-LargeA~\cite{lighttrack}	&55.0 &-	&55.2 &	&61.5 	&72.3	&- 		&- & -
\\
 \bottomrule
\end{tabular}}}
\caption{State-of-the-art comparison with lightweight trackers on LaSOT, GOT-10k, and NFS benchmarks. The best three are shown in \textbf{\textcolor{red}{red}}, \textbf{\textcolor{blue}{blue}}, and \textbf{\textcolor{cGreen}{green}} fonts.}
\label{tab:lightweight_sota}
\vspace{-2mm}
\end{center}
\end{table}

\subsection{Evaluation on Other Datasets}
We evaluate our tracker on several commonly used small-scale datasets, including VOT~\cite{vot2020, Kristan_2021_ICCV}, TNL2k~\cite{tnl2k}, NFS~\cite{NFS}, OTB2015~\cite{otb}, and UAV123~\cite{UAV}.
We also collect a number of state-of-the-art and baseline trackers for comparison. The results are shown in Tab.~\ref{tab-sota-small}.

{\noindent \textbf{VOT2021.}
\label{EvaluationVOT}
Using baseline experiments, we evaluate our tracker on the Visual Object Tracking Challenge (VOT2021)~\cite{vot2020}.
VOT consists of 60 challenging videos with mask annotation.
VOT2021 adopts expected average overlap (EAO) as the main metric, simultaneously considering the
trackers’ accuracy and robustness. 
We use the official evaluation tool and adopt AlphaRefine~\cite{AlphaRefine} to generate mask prediction.
As shown in Fig.~\ref{fig:eao2021}, our SuperSBT-Base variant yields the best performance among all methods.

{\noindent \textbf{NFS.}} We evaluate the proposed trackers on the 30 $fps$ version of the NFS~\cite{NFS} dataset. The NFS dataset contains challenging videos with fast-moving objects.
Our SuperSBT-Base variant achieves the best performance.

{\noindent \textbf{OTB2015.}} OTB2015~\cite{otb} contains 100 sequences and 11 challenge attributes.
Tab.~\ref{tab-sota-small} shows that our method does not achieve top performance on this dataset.
We note that OTB2015 is a small-scale dataset and is easily overfitted, and we did not deliberately tune the hyperparameters for this dataset.

{\noindent \textbf{UAV123.}}
UAV123~\cite{UAV} includes $123$ low-altitude aerial videos and adopts success and precision metrics for evaluation.
As shown in Tab.~\ref{tab-sota-small}, the SuperSBT-Base variant outperforms all other methods.

{\noindent \textbf{TNL2k.}}
TNL2k~\cite{tnl2k} is a recently released large-scale dataset with 700 challenging video sequences.
As shown in Tab.~\ref{tab-sota-small}, all of our three SuperSBT variants outperform all existing SOTA trackers.

\subsection{Comparing to Lightweight Trackers}
As shown in Tab.~\ref{tab:lightweight_sota}, compared to the specifically designed lightweight trackers (HiT~\cite{hit}, FEAR~\cite{fear}, E.T.Track~\cite{ettrack}, and LightTrack~\cite{lighttrack}), our SuperSBT-light/small tracker outperforms them by a great margin in three benchmarks, while still having comparable speed (over 100 FPS).
This indicates that our lightweight variants can achieve marvelous performance as well as maintain competitive efficiency.

\subsection{Ablation Study and Analysis}
\label{subsec:abla}

\begin{table}[t]
\centering

{
\centering
\begin{minipage}{1.0\linewidth}{\begin{center}
\tablestyle{5pt}{1.05}
\begin{tabular}{ccccc|c|c|c}
\Xhline{1.0pt}
No. & $N_{D}$  
& $N_{W}$
& Source
& Structure
& Parma. 
& Speed 
& AUC \\
\hline
1 & 17 & 320 & PVT~\cite{pvt} & Hier.
& 23M & 42 FPS &\textbf{57.5}  \\
2 & 23 & 320 & PVT~\cite{pvt}& Hier.
& 35M& 24 FPS &\textbf{58.2}  \\
\rowcolor{gray!20}  3 & 12 & 768 & ViT~\cite{vit} & plain 
& 85M & {102 FPS}& \textbf{61.1} \\
4 & {24} & {1024} & ViT~\cite{vit} & plain
& {305M} & 22 FPS& {\textbf{62.4} }\\
            \Xhline{1.0pt}

\end{tabular}
\end{center}}\end{minipage}

}
\caption{
Ablations on the effects of architecture variants: network depth $N_{D}$, width $N_{W}$, and model structure (plain/hierarchical). 
$N_{W}$ is the channel number of the main stage. Source refers to the type of ViT from which the tracking model is adapted.
The experiments are conducted on ITB~\cite{ifb}.
}
\label{tab:archi_factor} 
 \vspace{-2mm}
\end{table}


\begin{table}[t]
	\centering
    \setlength\tabcolsep{4pt}

       \resizebox{\linewidth}{!}{
    	\begin{tabular}{ccccc|cc|cc}
            \Xhline{1.0pt}
    		 \multirow{2}*{No.} & \multirow{2}*{Hierarchical} & \multirow{2}*{Pre-train} & \multirow{2}*{Temporal.}& \multirow{2}*{MixMLP.} & \multicolumn{2}{c|}{GOT-10k} & \multicolumn{2}{c}{LaSOT} \\
       
    	~ &	~ & ~ & ~  & ~ &AO &$\text{AO}_{50}$ & AUC & Prec. \\
      
            \Xhline{1.0pt}
          \rowcolor{gray!20}  1  & &   &   & & \textbf{62.1}  & \textbf{70.3}  & \textbf{55.7}  & \textbf{61.5} \\
             
          2 & \CheckmarkBold &   &  & & \textbf{65.9}  &\textbf{ 75.1}  & \textbf{58.2} & \textbf{64.6}  \\
            
         3  & &\CheckmarkBold  &  & &\textbf{71.0}  & \textbf{81.4}  &\textbf{66.5}  &\textbf{71.0}  \\
            
         4 & &  & \CheckmarkBold &  & \textbf{64.2}  & \textbf{71.5}  &\textbf{57.1}  & \textbf{62.4}  \\
        
        5 & &  &  & \CheckmarkBold &\textbf{63.5} & \textbf{71.8}  &\textbf{57.8}  & \textbf{63.9}  \\

         6  &     \CheckmarkBold & \CheckmarkBold  &\CheckmarkBold & \CheckmarkBold &\textbf{74.2}  & \textbf{83.6} &\textbf{69.9} & \textbf{76.0}\\
            \Xhline{1.0pt}
    	\end{tabular}
    }
       \caption{Improvements of SuperSBT. \CheckmarkBold denotes adopting hierarchical structure, masked image modeling pre-training, temporal modeling, and Mix-MLP head.}
        \vspace{-2mm}
    \label{tab:ablation_supersbt}
\end{table}

\begin{figure}[t]
	\centering{\includegraphics[scale = 0.55]{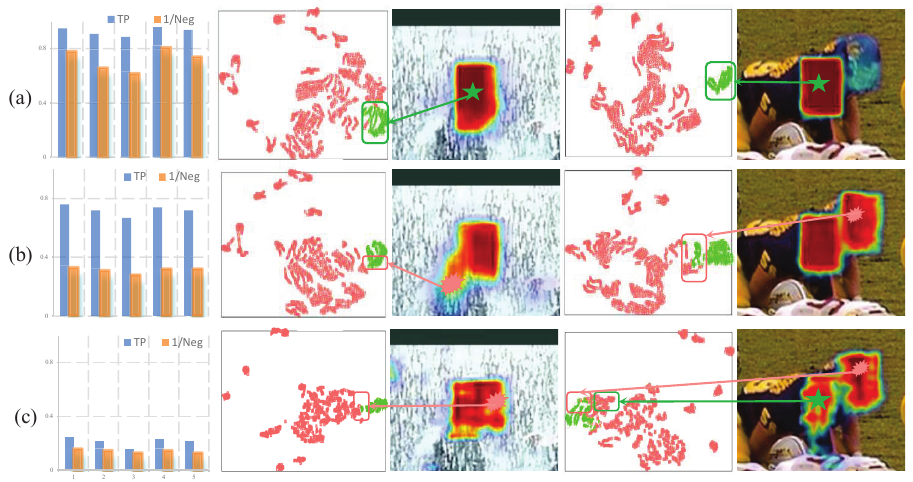}} 
	\caption{(a), (b), (c) denote three trackers (refer to Sec.~\ref{sec:discri}). The first sub-figure indicates the average true positive rate and average negative numbers of negative objects. The other sub-figures denote the T-SNE and classification maps.}
 \vspace{-2mm}
	\label{fig:aba_spatial}
\end{figure}

\begin{figure}[t]
	\centering{\includegraphics[scale = 0.46]{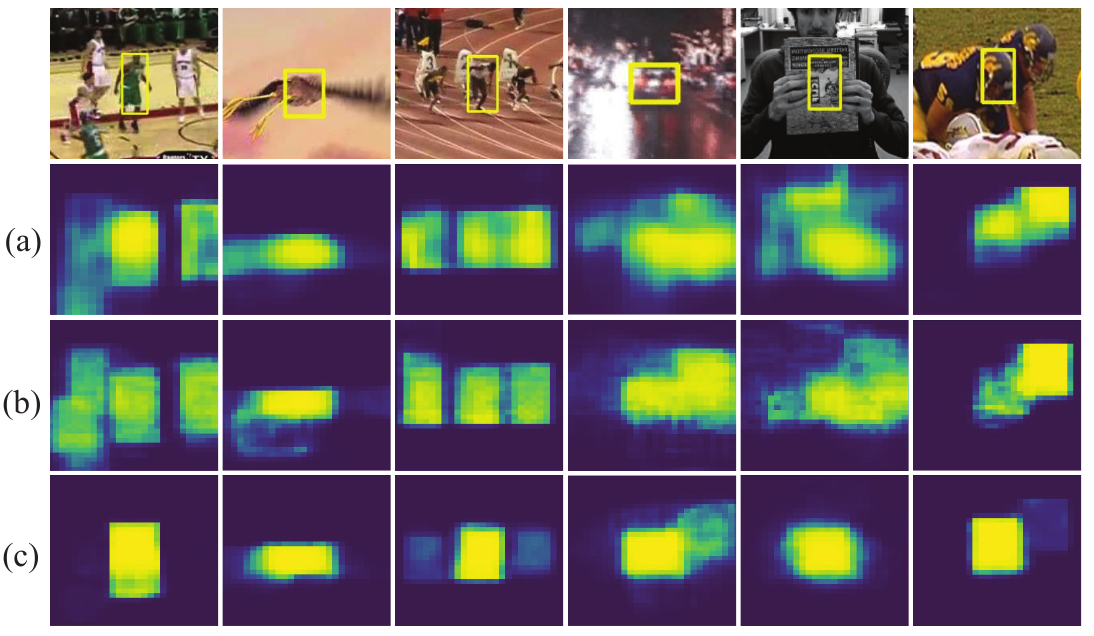}} 
	\caption{Visualization of classification (Cls) map on Hi-SBT tracker with three different settings. (a) layer-wise aggregation with DW-Corr; (b) layer-wise aggregation with FRM-CA block; (c) correlation-embedded. }
	\label{fig:vis_sea}

\end{figure}

\begin{figure}[!t]
	\centering{\includegraphics[scale = 0.26]{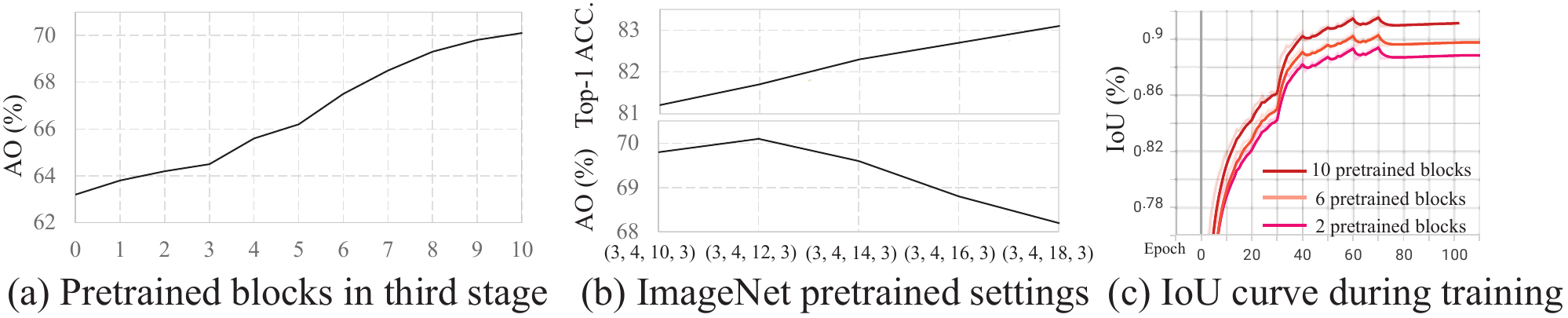}} 
	\caption{Tracking performance of SBT with various pre-trained settings. (a) pre-trained layer number. (b) different models are used to initialize the tracking model. (c) the IoU curves during training. }
	\label{fig:aba_imagenet}
  \vspace{-2mm}
\end{figure}

\noindent\textbf{Single-branch structure.}
As presented in Tab.~\ref{tab:ablation_structure}, the correlation-embedded SBT (\ding{178}, \ding{179}, \ding{180}) leads to a significant enhancement in tracking performance for all correlation cases (\ding{175}, \ding{176}, \ding{177}).%
Comparing to layer-wise aggregation, correlation-embedded trackers outperform their CNN-based or attention-based counterparts ($65.9\%$ of \ding{179} vs. $60.1\%$ of \ding{175}, $65.0\%$ of \ding{178} vs. $61.5\%$ of \ding{176}, $39.2\%$ of \ding{180} vs. $30.3\%$ of \ding{174}). 
The empirical evidence demonstrates that utilizing multi-level features is more effective under the structure of SBT.
It also verifies that CA works better than DW-Corr in feature correlation ($60.1\%$ of \ding{175} vs. $61.5\%$ of \ding{176}). 
Fig.~\ref{fig:vis_sea} also manifests the superiority of correlation-embedded structure.

\noindent\textbf{Target-aware feature embedding.}
\label{sec:discri}
We explore the features of three different settings in two folds: maintaining spatial location information and identifying the target from distractor objects.
To begin with, we train three models that have only Cls head for localizing the target: 
(a) Correlation-embedded tracker. 
(b) Siamese correlation with SBT.
(c) Siamese correlation with ResNet-50.  
We randomly jitter the search image around the target in the five hardest videos from the OTB~\cite{otb} benchmark. We evaluate only the Cls map for localization.%
In Fig.~\ref{fig:aba_spatial}, we observe that the true positive rate of the target ground truth demonstrates that (a) and (b) can preserve more spatial information compared to (c) CNN. The T-SNE/Cls map also exhibits the target-dependent characteristic of (a) features. The average negative objects (largest connected components) of (a) are higher than (b), indicating the effectiveness of the correlation embedding.

\noindent\textbf{Fast training convergence.}
Our tracking model, except for its prediction heads, can benefit directly from pre-trained weights on ImageNet, which is different from existing trackers~\cite{transt, stark, dualtfr}.  
A strong correlation exists between tracking performance and the number of pre-trained blocks, as depicted in Fig.~\ref{fig:aba_imagenet}(a).%
We also investigate the impacts of SBT model variants.  
In Fig.~\ref{fig:aba_imagenet}(b), the SBT tracking model prefers consistent block numbers for pre-training. We also observe that SBT converges faster, and the stabilized IoU value rises with more pre-trained model weights.

\noindent\textbf{Effect of architectural variants.}
We investigate the effects of three architectural factors on the performance and inference speed of our method, namely the number of transformer blocks $N_{D}$ (depth), the channel dimension in each block $N_{W}$ (width), and the model structure (isotropic/hierarchical).  
Tab.~\ref{tab:archi_factor} shows that model performance depends very weakly on its shape (depth and width) but is significantly influenced by its scale (Models No.3/No.4 exhibit notably better performance than small-scale models).  
The observation that model No.1 has a smaller scale but lower speed highlights that the speed heavily hinges on the model shape.

{\noindent \textbf{Improvements of SuperSBT.}}
In Tab.~\ref{tab:ablation_supersbt}, we study the effects of three modifications made to SuperSBT: hierarchical structure, masked image pre-training, Mix-MLP head, and temporal modeling scheme. 
We observe that SuperSBT significantly outperforms the baseline, confirming the efficacy of these three modifications.
Moreover, consistent performance gains can be obtained when each of the five modifications is applied individually, with the largest improvement yielded by masked image pre-training.

\section{Conclusion}
 \label{sec:conclusion}
 In this work, we propose a novel 
fully transformer-based Single-Branch Tracking framework (SBT).
 Our SBT greatly simplifies the popular Siamese tracking pipeline by unifying feature extraction and correlation steps as one stage.
 Moreover, we conduct a systematic study on SBT tracking and summarize a bunch of effective design principles. 
Based on the summarized principles, we have developed an improved SuperSBT model for tracking. 
Enhanced by reasonable architecture variant design, masked image modeling pre-training, and temporal modeling, 
our SuperSBT model delivers superior results while raising the running speed with an even larger model capacity. 
Experiments on eight VOT benchmarks verify that our SBT and SuperSBT variants achieve SOTA results while maintaining a simple architecture, showing great potential to serve as a strong baseline tracker.



\bibliographystyle{IEEEtran}
\bibliography{egbib}

\begin{IEEEbiography}[{\includegraphics[width=1in,height=1.25in,clip,keepaspectratio]{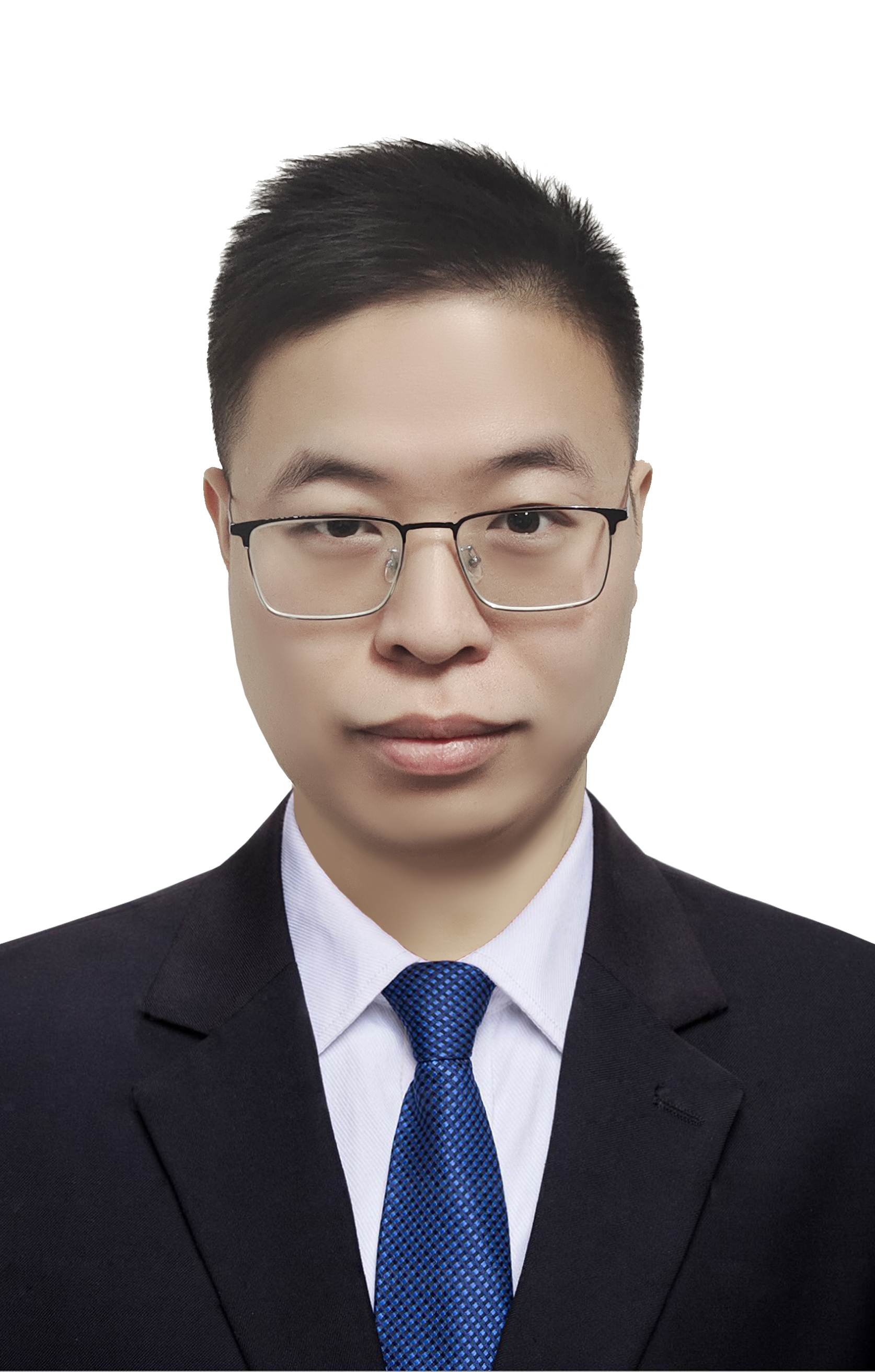}}]
{Fei Xie} is currently a Ph.D. student at Shanghai Jiao Tong University from the year 2022. 
He obtained a Master's degree from Southeast University in China in 2019 and a Bachelor's degree from Huazhong University of Science and Technology. 
He also worked as a research intern in Microsoft Research Aisa from April 2021 to June 2023. 
His research interests include computer vision and deep learning, especially in object detection, video tracking, and visual backbone networks.  
\end{IEEEbiography}

\begin{IEEEbiography}[{\includegraphics[width=1in,height=1.25in,clip,keepaspectratio]{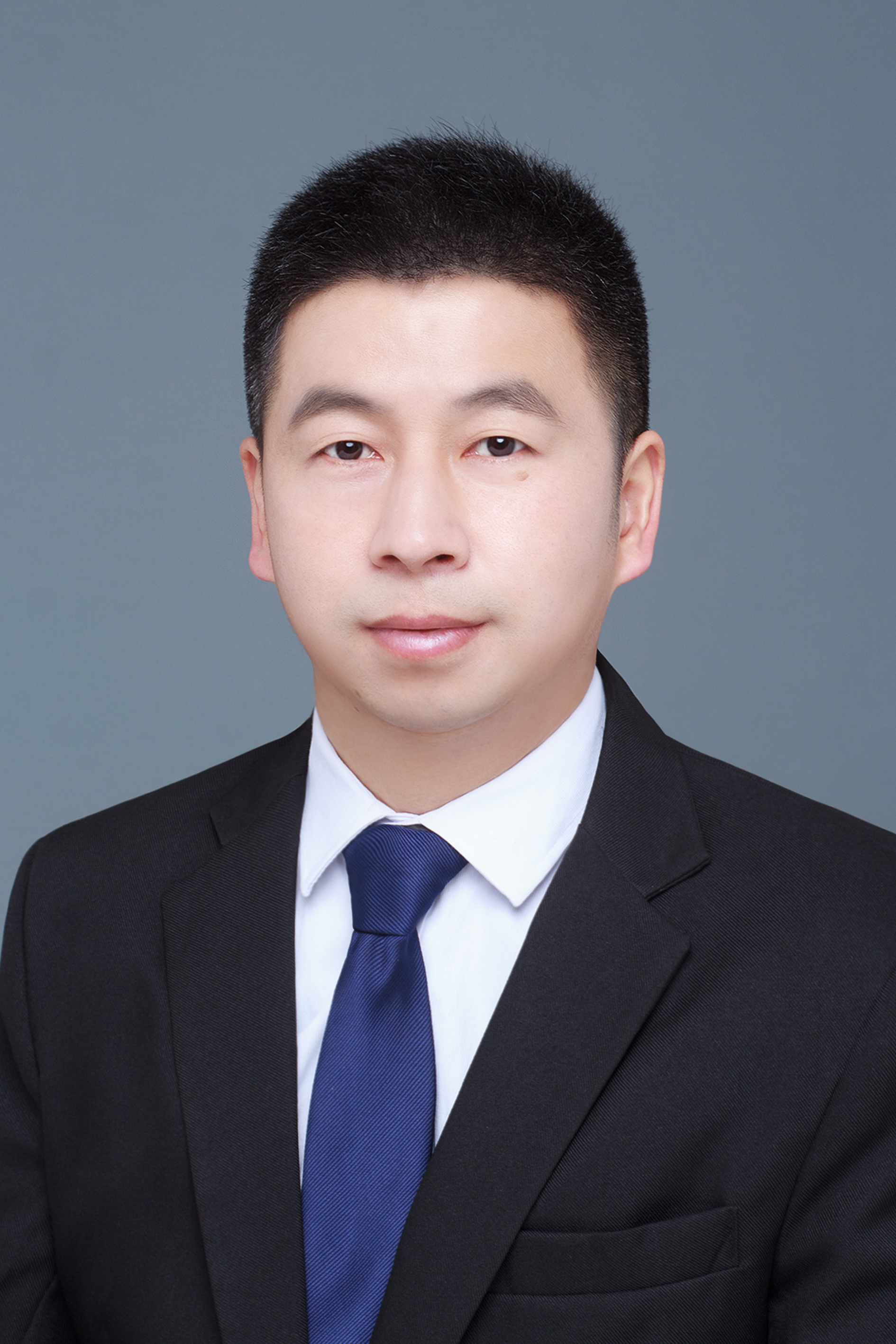}}]{Wankou Yang} received the B.S., M.S. and Ph.D. degrees in the School of Computer Science and Technology, Nanjing University of Science and Technology (NUST), China, respectively in 2002, 2004, and 2009. From July 2009 to Aug.2011, he worked as a Postdoctoral Fellow in the School of Automation, Southeast University, China. He is the professor
in the School of Automation, at Southeast University. His research interests include pattern recognition and computer vision.
\end{IEEEbiography}

\begin{IEEEbiography}[{\includegraphics[width=1in,height=1.25in,clip,keepaspectratio]{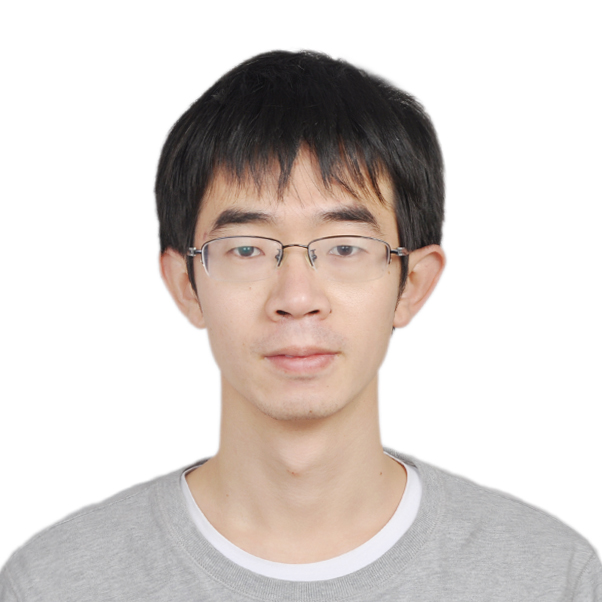}}]{Chunyu Wang} received the PhD degree in computer science from Peking University, in 2016. He is a senior researcher with Microsoft Research Asia. His research interests include computer vision and machine learning algorithms, and their applications to solve real world problems.
\end{IEEEbiography}

\begin{IEEEbiography}[{\includegraphics[width=1in,height=1.25in,clip,keepaspectratio]{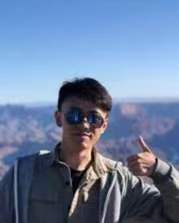}}]{Lei Chu} received his Ph.D. degree in computer science from The University of Hong Kong in 2020. Presently, he holds the position of Senior Researcher at Microsoft Research Asia, within the Media Computing Group. His research is primarily focused on 3D reconstruction, scene understanding, and video understanding.

\end{IEEEbiography}

\begin{IEEEbiography}[{\includegraphics[width=1in,height=1.25in,clip,keepaspectratio]{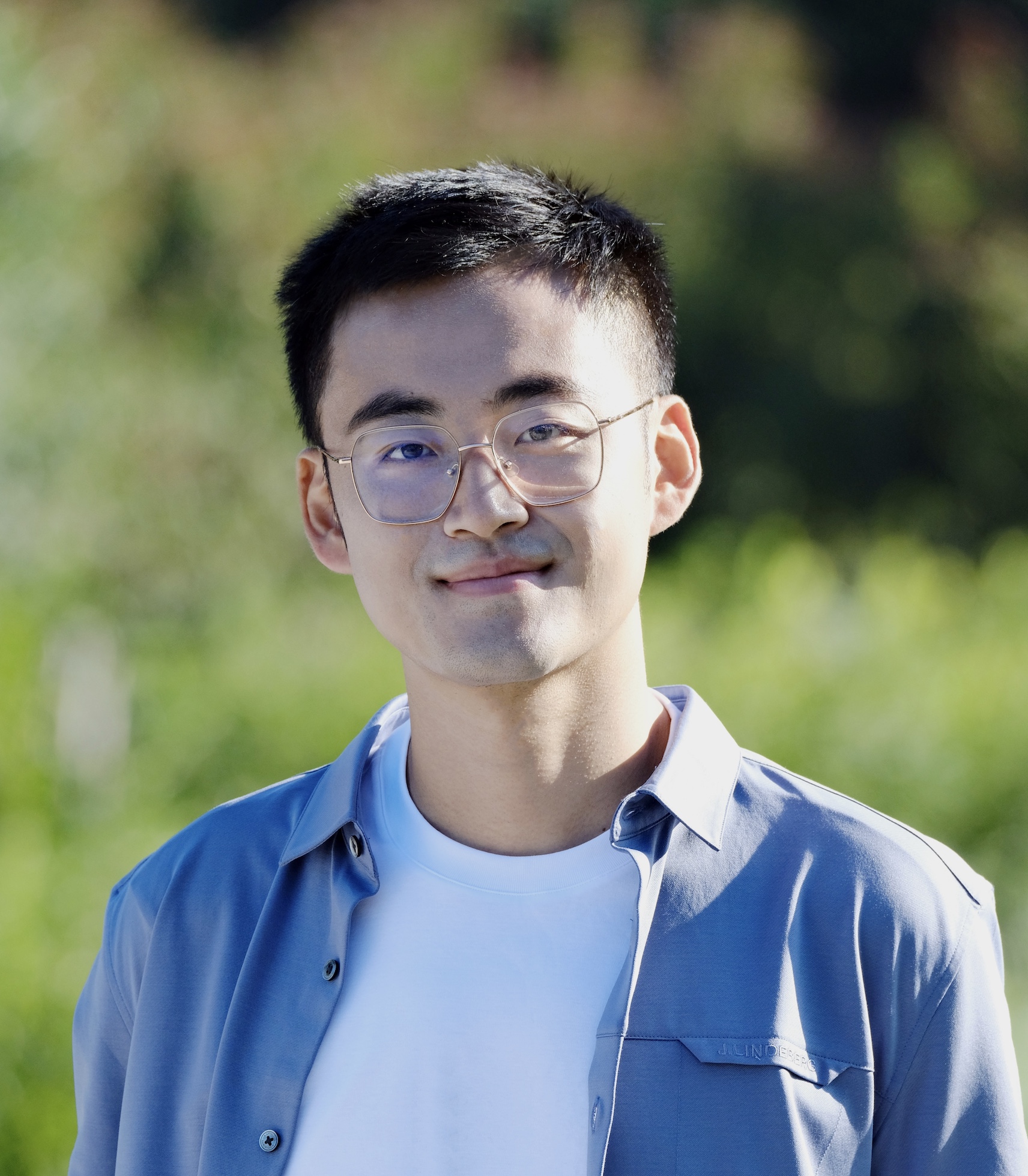}}]{Yue Cao}received the BE degree in computer software and PhD degree in software engineering from Tsinghua University, China. He was awarded the Top-Grade Scholarship of Tsinghua University
in 2018. He was a researcher with Microsoft Research Asia. 
His work of Swin Transformer won the Best Paper Award (Marr Prize) of ICCV 2021. 
His research interests include computer vision and deep learning, especially in self-attention modeling.
\end{IEEEbiography}

\begin{IEEEbiography}
[{\includegraphics[width=1in,clip,keepaspectratio]{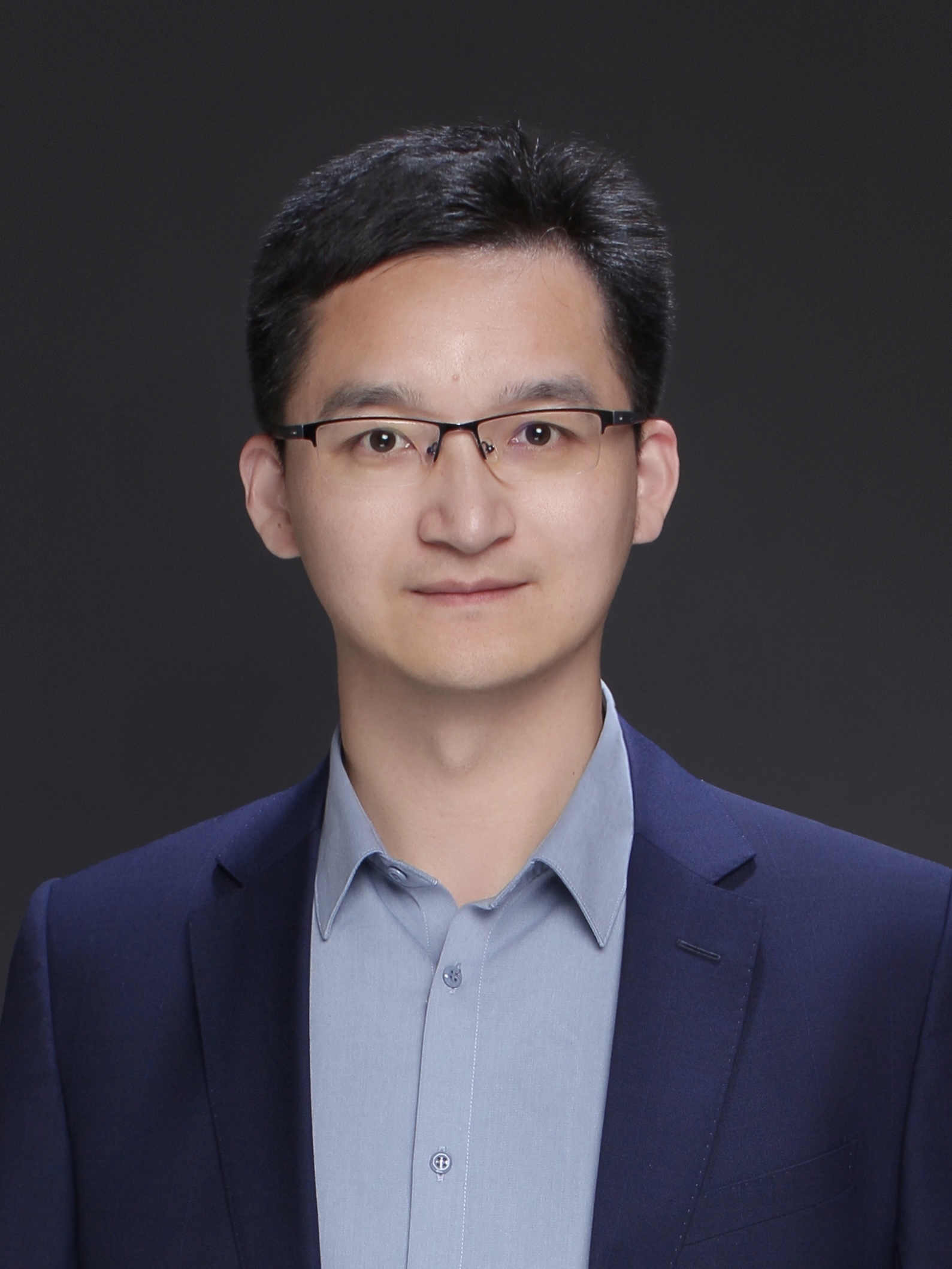}}]{Chao Ma} is a professor of computer science at Shanghai Jiao Tong University, China. His research interests include computer vision and machine learning. Prior to joining Shanghai Jiao Tong University, he was a senior research associate with the School of Computer Science at The University of Adelaide. He received his Ph.D. degree from Shanghai Jiao Tong University in 2016. From the fall of 2013 to the fall of 2015, he was sponsored by the China Scholarship Council as a visiting Ph.D. student at the University of California at Merced. 
\end{IEEEbiography}

\begin{IEEEbiography}[{\includegraphics[width=1in,height=1.25in,clip,keepaspectratio]{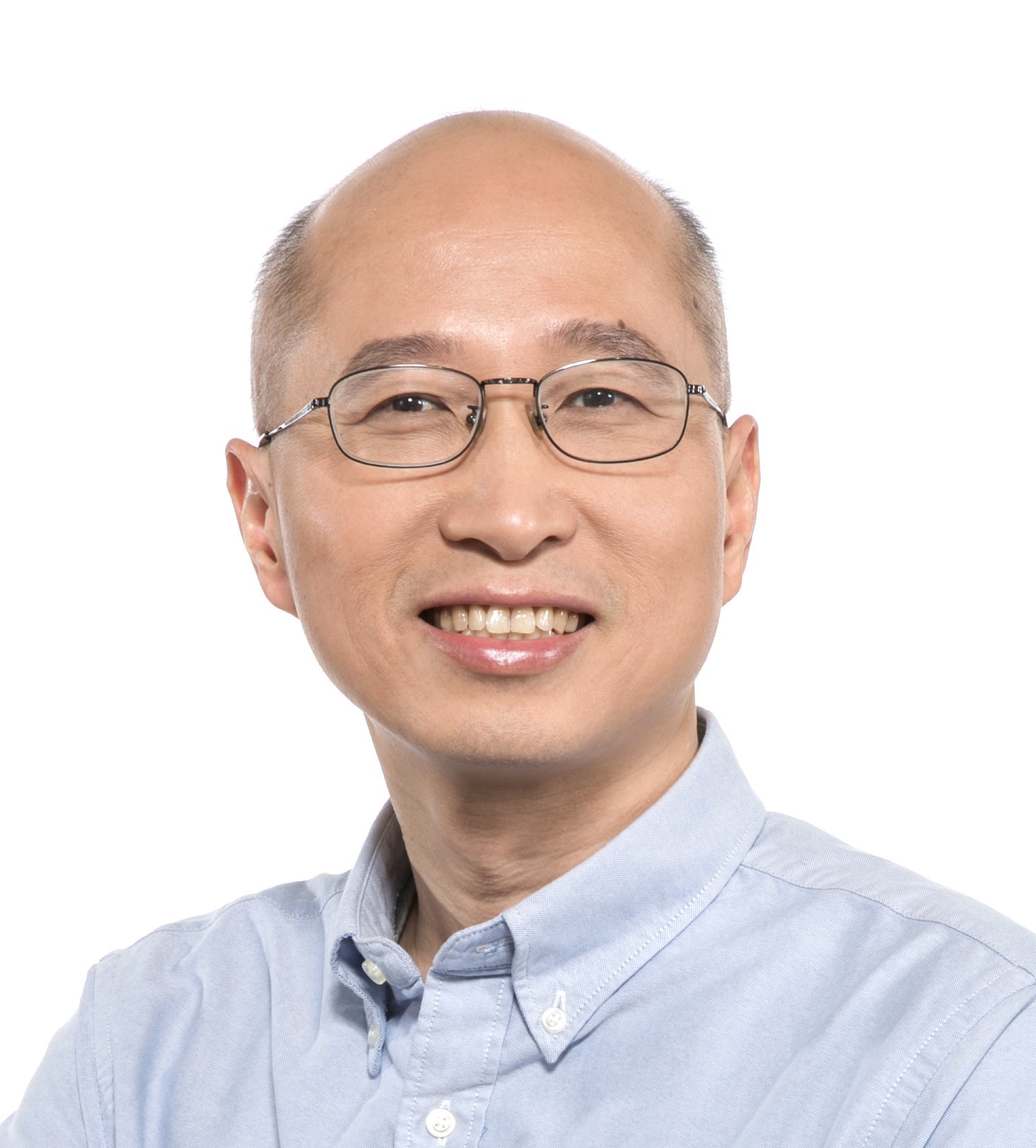}}]{Wenjun Zeng} Wenjun Zeng (Fellow, IEEE) received the B.E. degree from Tsinghua University, Beijing, China, in 1990, the M.S. degree from the University of Notre Dame, Notre Dame, IN, USA, in 1993, and the Ph.D. degree from Princeton University, Princeton, NJ, USA, in 1997. He has been a Chair Professor and the Vice President for Research at the Eastern Institute for Advanced Study (EIAS) / Eastern Institute of Technology (EIT), Ningbo, China, since October 2021. He is also the founding Executive Director of the Ningbo Institute of Digital Twin. He was a Sr. Principal Research Manager and a member of the Senior Leadership Team at Microsoft Research Asia, Beijing, from 2014 to 2021, where he led the video analytics research empowering the Microsoft Cognitive Services, Azure Media Analytics Services, Office, and Windows Machine Learning. He was with University of Missouri, Columbia, MO, USA from 2003 to 2016, most recently as a Full Professor. Prior to that, he had worked for PacketVideo Corp., Sharp Labs of America, Bell Labs, and Panasonic Technology. He has contributed significantly to the development of international standards (ISO MPEG, JPEG2000, and OMA). Dr. Zeng is on the Editorial Board of the International Journal of Computer Vision. He was an Associate Editor-in-Chief of the IEEE Multimedia Magazine and an Associate Editor of the IEEE TRANSACTIONS ON CIRCUITS AND SYSTEMS FOR VIDEO TECHNOLOGY, IEEE TRANSACTIONS ON INFORMATION FORENSICS AND SECURITY, and IEEE TRANSACTIONS ON MULTIMEDIA (TMM). He was on the Steering Committee of IEEE TRANSACTIONS ON MOBILE COMPUTING and IEEE TMM. He served as the Steering Committee Chair of IEEE ICME in 2010 and 2011, and has served as the General Chair or TPC Chair for several IEEE conferences (e.g., ICME’2018, ICIP’2017). He was the recipient of several best paper awards.
\end{IEEEbiography}

\vfill

\end{document}